\documentclass[5p]{elsarticle}%

\usepackage[super]{nth}
\usepackage{amssymb}
\usepackage{lineno} 
\usepackage{amsmath}
\usepackage[dvipsnames]{xcolor}
\usepackage{bm}
\usepackage{hyperref}
\usepackage{adjustbox}
\usepackage{enumitem}
\usepackage[normalem]{ulem}
\usepackage{subcaption}
\usepackage{array,multirow,graphicx}
\usepackage{xurl}
\usepackage{breakcites}
\usepackage{microtype}
\usepackage{soul}

\DeclareMathOperator*{\argmax}{arg\,max\,}
\DeclareMathOperator*{\argmin}{arg\,min\,}
\graphicspath{{images/}}
\definecolor{cyan}{RGB}{191, 255, 255}
\definecolor{lime}{RGB}{191, 255, 191}
\definecolor{yellow}{RGB}{255, 255, 191}
\definecolor{reddish}{RGB}{255, 192, 191}
\definecolor{purple}{RGB}{255, 191, 255}
\definecolor{plt_C2}{RGB}{44, 160, 44}

\makeatletter
\def\ps@pprintTitle{%
 \let\@oddhead\@empty
 \let\@evenhead\@empty
 \def\@oddfoot{}%
 \let\@evenfoot\@oddfoot}
\makeatother

\begin{document}


\begin{frontmatter}



\title{Country-wide Retrieval of Forest Structure From \\ Optical and SAR Satellite Imagery With Deep Ensembles}

\author[inst1]{Alexander Becker\corref{cor1}}
\cortext[cor1]{Corresponding author; \href{mailto:alexander.becker@geod.baug.ethz.ch}{\texttt{alexander.becker@geod.baug.ethz.ch}}}
\author[inst1]{Stefania Russo}
\author[inst2]{Stefano Puliti}
\author[inst1]{Nico Lang}
\author[inst1]{\\Konrad Schindler}
\author[inst1,inst3]{Jan Dirk Wegner}

\affiliation[inst1]{organization={EcoVision Lab, Photogrammetry and Remote Sensing, ETH Zurich}, country={Switzerland}}

\affiliation[inst2]{organization={National Forest Inventory Department, Norwegian Institute of Bioeconomy Research (NIBIO)}, country={Norway}}
            
\affiliation[inst3]{organization={Institute for Computational Science, University of Zurich}, country={Switzerland}}

\begin{abstract}
\begin{linenumbers}
Monitoring and managing Earth's forests in an informed manner is an important requirement for addressing challenges like biodiversity loss and climate change. While traditional in situ or aerial campaigns for forest assessments provide accurate data for analysis at regional level, scaling them to entire countries and beyond with high temporal resolution is hardly possible. In this work, we propose a method based on deep ensembles that densely estimates forest structure variables at country-scale with 10-meter resolution, using freely available satellite imagery as input. Our method jointly transforms Sentinel-2 optical images and Sentinel-1 synthetic-aperture radar images into maps of five different forest structure variables: \nth{95} height percentile, mean height, density, Gini coefficient, and fractional cover. We train and test our model on reference data from 41 airborne laser scanning missions across Norway and demonstrate that it is able to generalize to unseen test regions, achieving normalized mean absolute errors between 11\% and 15\%, depending on the variable. Our work is also the first to propose a variant of so-called Bayesian deep learning to densely predict multiple forest structure variables with well-calibrated uncertainty estimates from satellite imagery. The uncertainty information increases the trustworthiness of the model and its suitability for downstream tasks that require reliable confidence estimates as a basis for decision making. We present an extensive set of experiments to validate the accuracy of the predicted maps as well as the quality of the predicted uncertainties. To demonstrate scalability, we provide Norway-wide maps for the five forest structure variables.
\end{linenumbers}
\end{abstract}



\end{frontmatter}

\section{Introduction}
\label{sec:introduction}
Forest structure relates to the three-dimensional (3D) spatial arrangement of the plant community within a forest and is both a result and a driver of ecosystem processes and biological diversity \citep{spies1998}. Forest structure is strongly correlated with a forest’s ability to store carbon and to provide habitat for a variety of species \citep{turner2003remote,bergen2009remote,dubayah2020global}. Given the large extent and dynamic nature of forest ecosystems, it is important to develop ways of mapping and monitoring their structure consistently through space and time \citep{VALBUENA2020}. Such maps can support a more informed and dynamic management of forest resources, required to tackle important global challenges such as the loss of biodiversity; and the mitigation of, as well as adaptation to climate change.\\

Measuring forest structure has long been an expensive and time-consuming task. Traditional in-situ methods are limited to the recording of a few tree characteristics measurable from the ground (e.g., diameter at breast height, tree species, to some degree tree height), and restricted to small sample plots sparsely scattered across the landscape. Consequently, in-situ measurements are not sufficient for a spatially explicit understanding of forest structures. To collect additional, explicit forest structure measurements, field direct observations can be complemented with terrestrial laser scanning (TLS). Still, TLS in practice only yields local samples, as it has a limited range and is affected by occlusions \citep{calders2020terrestrial}. In contrast, airborne laser scanning (ALS) allows one to densely observe forest structure at regional scales. Forest structural variables derived from ALS, such as canopy height, cover, or density, have long been used for the wall-to-wall characterization of single-tree \citep{Hyyppa2001} and forest biophysical properties \citep{Nilsson1996, Naesset2002}. But the high operational cost, which scales more or less linearly with flight time, limits the coverage and revisit time of ALS campaigns and is a bottleneck for country-scale applications.\\

\begin{figure*}[!ht]
	\centering
    \vspace{-5px}
	\includegraphics[width=\linewidth]{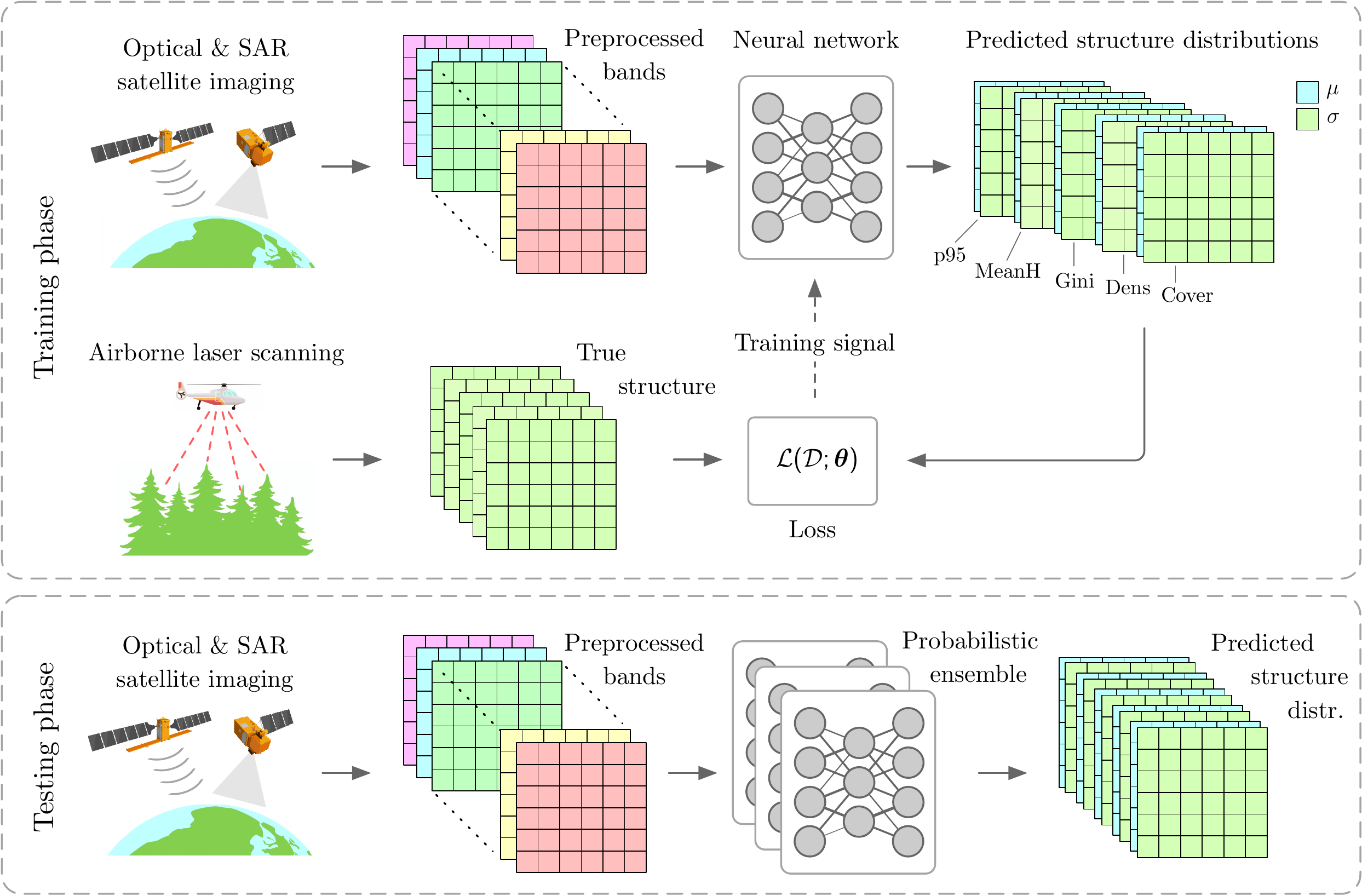}
	\caption{High-level overview of our method. \emph{Top panel}: During training, the parameters of a neural network model are optimized to reduce the deviation between predicted and ALS-derived forest structure variables. \emph{Bottom panel}: During inference, an \emph{ensemble} of multiple neural networks predicts a distribution over forest structure variables, given only optical and SAR satellite images with 10$\,$m GSD as input. Satellite icons from \citep{satellite_icons}.}
	\label{fig:highlevel}
\end{figure*}

The natural next step to scale beyond local and single-date ALS campaigns is to take advantage of the wealth of open data from Earth observation satellites. A prominent example is the European Union's Copernicus program, with multiple satellite missions whose data products are useful to predict forest structure dynamics \citep{Puliti_change2021}. In particular, the Sentinel-1 (synthetic-aperture radar, SAR) and Sentinel-2 (multispectral optical) missions provide near-global coverage, with both high spatial resolution ($\sim$10$\,$m ground sampling distance, GSD) and frequent revisit times ($\leq$5$\,$days; the effective temporal resolution of the optical sensor is reduced by the cloud coverage). 
These missions deliver observations over a wide spectral range at an unprecedented rate, but can they be converted to forest structure products? A new challenge arises to retrieve the relevant information from multispectral image data at $\sim$10$\,$m GSD, rather than LiDAR waveforms with \textless1$\,$m\textsuperscript{2} footprints. In the absence of a tractable radiative transfer model, supervised machine learning offers a statistical approach to retrieve environmental parameters from image data. In recent years, deep neural networks (DNNs) have emerged as the mainstream tool for image analysis, including their applications in remote sensing and, in particular, mapping of forest structure \citep{Lang2019}. DNNs have brought substantial performance gains through their ability to learn the complete functional mapping from raw image values to the desired output variables. 
While DNNs often achieve high precision, their outputs are typically limited to point estimates and/or are overly confident\citep{Guo2017, gast2018lightweight}. As well-calibrated estimates of uncertainty are important whenever the model outputs feed into critical decisions or probabilistic models, the Bayesian interpretation of model outputs is an active line of research \citep{Blundell2015, Gal2016, Guo2017, Kendall2017, Lakshminarayanan2017, Ovadia2019, Gustafsson2020, Wilson2020}.
\\

In this paper, we go a step further and exploit recent advances from the field of \emph{Bayesian deep learning} (BDL) to densely map forest structure variables \emph{and} associated {uncertainties} from optical and SAR satellite images.
BDL is understood as the deep learning counterpart to traditional Bayesian inference techniques, for instance those for linear regression \citep[Ch.~2 to 2.1.1]{Rasmussen2004} or Gaussian process regression \citep[Ch.~2.1.2 to 2.3]{Rasmussen2004}. Instead of committing to a single solution of model parameters, BDL is characterized by (approximate) marginalization over a posterior distribution of all possible models, given some assumptions about the prior and the data likelihood.
Such a principled approach is particularly attractive for data-driven models, where the model uncertainty is not a consequence of explicit modeling decisions, but instead is caused by the limited training data. Bayesian marginalization may improve the predictive accuracy, but more importantly it can give the user access to well-calibrated estimates of the predictive uncertainty, which take into consideration how well a given model prediction is supported by training data. 
This ability to self-diagnose the reliability of each individual prediction is indispensable for many downstream tasks that ingest the model output as their input. As an example, when making informed decisions, it is crucial to know how reliable the data are that the decisions are based on. Uncertainty estimates have been successfully used for decision making in many fields of application \citep{NRC06, Soroudi13, Martin19, Sniazhko19, Alzate21}, and we argue that they are equally relevant in forest management in order to optimize decisions for cost, carbon stock, biodiversity etc.

Our method (see overview in Fig.~\ref{fig:highlevel}) is fully supervised and consists of a training and a testing phase. During the training phase, a deep convolutional neural network is optimized on training data, i.e., optical and SAR images with associated, pixel-wise forest structure reference values computed from ALS data. The fitting employs a loss function that does not just penalize deviations between model outputs and ALS reference data. Instead, the loss is proportional to the negative log posterior probability over the model parameters given the data, assuming a zero-mean normally distributed prior (acting as a regularizer) and a Gaussian data likelihood. Thus, for each pixel and variable, the model outputs the parameters of a Gaussian distribution, thereby capturing uncertainty inherent in the input data (aleatoric uncertainty). During the testing phase, multiple trained networks are combined into an ensemble model, exploiting the stochastic nature of neural network initialization and training. The predicted distributions of all networks are aggregated into an ensemble estimate of the distribution over structure variables. The aggregation of predictions from independently trained models corresponds to a sample-based approximation of Bayesian marginalization \citep{Gustafsson2020, Wilson2020}. It captures the uncertainty in the model parameters (epistemic uncertainty), which is combined with the aleatoric uncertainty to obtain the overall uncertainty of the predicted forest structure variables.
\\

In this work, we propose a deep ensembles approach to predict structural forest variables often used in modeling forest biophysical properties (\nth{95} height percentile, mean height, density, Gini coefficient and fractional cover), using optical and SAR satellite images as input. 
Our approach can be seen as a scalable complement to regional ALS, enabling forest structure mapping with 10-meter ground sampling distance (GSD) at country-scale and with a high update frequency (in the extreme case down to five days), at a low cost.
Moreover, the estimated forest structure maps come with an individual, calibrated uncertainty estimate for every structure variable at every single pixel.
We conduct extensive experiments on a test region in Norway, for which reference data from a full-waveform ALS campaign is available. We then apply our method to compute a country-wide forest structure map for all of Norway, that we make available online.\footnote{\url{https://albecker.users.earthengine.app/view/forest}} Source code for training and testing the model is provided, too.\footnote{\url{https://github.com/prs-eth/bayes-forest-structure}} 

\section{Related Work}
\label{sec:related_work}
\subsection{Remote sensing of structural forest variables}
 Since its early days, ALS research has seen a tremendous growth, and forest structural variables derived from ALS are broadly used as a source of auxiliary information for the large scale characterization of forest ecosystems. Many different types of modeling techniques have been tested over the years, most of which rely on linking field measured forest biophysical properties (e.g. biomass) with forest structural predictor variables derived from ALS, like height percentiles, vegetation density, forest cover, or foliage height diversity \citep{Naesset2004laser, Naesset2007}.
 More recently, there has been a trend to directly use ALS-derived structural variables to map forest structural \citep{COOPS2016,Valbuena2017, Adnan2021} and functional diversity \citep{schneider2017mapping, ZHENG2021}. While ALS data are the most detailed source of information to characterize forest structure, their geographic and temporal availability remains limited.
 
 If the goal is frequent, or even continuous, monitoring of large forest areas, a more promising data source are freely available satellite images, such as those provided by the Landsat and Copernicus programs. Landsat-based approaches have mostly been relying on time series features to map forest structure \citep{tyukavina2015aboveground,hansen2016mapping,potapov2019annual, potapov2021mapping}. Also  Copernicus' Sentinel-1 and Sentinel-2 missions, with 10~m GSD and less than five days revisit time, offer dense time series of observations that are suitable for forest monitoring. Previous studies have demonstrated the usefulness of Sentinel data to map and estimate key forest biophysical variables (e.g., above ground biomass) \citep{Laurin2018, Puliti_ArcticDEM_2020, Breidenbach2021SPECIES} and their dynamics trough time \citep{Puliti_change2021}. In the past two years, a growing body of literature has shown the possibility to map forest canopy height with Sentinel-2 \citep{Lang2019, Shimizu2020, Astola2021, lang2021high}. Since canopy height is only one of many ecosystem characteristics that can be obtained from ALS, our work aims to understand whether Sentinel data can provide maps for a more comprehensive spectrum of structural variables, like vegetation density, cover, and complexity. Like~\cite{Lang2019} our model learns to extract predictive texture features from single input images.

\subsection{Deep learning in remote sensing}
In the last decade, deep learning has revolutionized the way information is extracted from images. In particular, convolutional neural networks (CNNs) have achieved unprecedented results in areas like image classification \citep{Krizhevsky2012, Simonyan2014, He2015}, semantic segmentation \citep{Long2015, Chen2016}, object detection \citep{Szegedy2013, Girshick2014, Redmon2015} and further perception tasks.

In addition, deep learning is increasingly being applied to vision tasks in remote sensing, such as super resolution \citep{lanaras2018super, deLutio2019} or change detection \citep{daudt2018}. Traditional applications include land cover classification from aerial \citep{Kaiser2017,Marmanis2018,zhang2019joint} or satellite images \citep{helber2019eurosat}. Agricultural crop type classification, a specific type of land cover, is a well-studied task, where most authors exploit features from time-series data \citep{russwurm2018multi, m2019semantic, garnot2019time, russwurm2020self, turkoglu2021crop, turkoglu2021gating}.
More related to our work are methods that seek to predict biophysical indicators such as crop yield from climate data and Enhanced Vegetation Index maps \citep{Kuwata2015}, the prediction of sea ice concentration from RADARSAT SAR satellite images \citep{Wang2016} or tree density estimation from Sentinel-2 optical images \citep{Rodriguez2018, rodriguez2021mapping}.
Most works so far do not accompany their map products with calibrated uncertainty estimates. Exceptions include \citep{rodriguez2021mapping}, where uncertainties are used to guide active learning; and \citep{lang2021global}, where BDL is integrated into a 1-dimensional CNN for spaceborne LiDAR analysis.
The prior work most relevant for our paper is \cite{Lang2019}, where a CNN was developed to predict country-wide vegetation height maps from Sentinel-2 optical images. Their model was trained and evaluated for Gabon and Switzerland, where training data were derived from LiDAR measurements and photogrammetric surface reconstruction, respectively. 
Here, we predict five forest structural variables jointly and estimate their predictive uncertainties using an ensemble of probabilistic neural networks. Moreover, we explore the potential of data fusion using Sentinel-1 SAR images as an additional input signal on top of the Sentinel-2 optical bands.

\subsection{Uncertainty in deep learning}
\label{sec:rel_uncertainty}
Although deep learning has become the most widely used method in computer vision, model outputs are often trusted ``blindly". In regression problems, outputs are usually point estimates with no attached notion of uncertainty, while classification scores have been shown to be overconfident \citep{Guo2017, Lakshminarayanan2017}. To mitigate this effect and to develop more trustworthy models, reliably quantifying the predictive uncertainty is important -- and an open research question \citep{Gustafsson2020}.

Usually, uncertainty in machine learning is decomposed into \emph{aleatoric} and \emph{epistemic} uncertainty \citep{Gal2016}. The former is assumed to be inherent in the observations -- e.g. resulting from sensor noise or lack of signal -- and can thus not be ``explained away" with more training data. On the contrary, epistemic uncertainty captures uncertainty in the model itself, i.e., the lack of knowledge due to not having seen enough training data for the respective region of the input space. Unlike aleatoric uncertainty, it cannot simply be learned from data -- instead, it generally requires marginalization over an (approximate) posterior distribution over model parameters. These methods are colloquially referred to as \emph{Bayesian deep learning} \citep{Kendall2017}; common approaches include variational Bayesian inference \citep{Ranganath2013, Blundell2015} and methods based on Markov chain Monte Carlo (MCMC) sampling \citep{Neal1996, Welling2011, Chen2014}. In this work, we use a deep ensemble \citep{Lakshminarayanan2017}, a method specifically developed for deep neural networks that can be understood to perform approximate Bayesian inference \citep{Gustafsson2020}. The general idea is to train an ensemble of $M$ independent models on the same data, each initialized with a different set of random weights. The randomness inherent in the weight initialization, as well as random sampling of training batches, causes each model to converge to a different local minimum in the solution space, and the resulting weights can be interpreted as samples from an approximate posterior distribution \citep{Gustafsson2020,Wilson2020}. In practice, among all methods performing approximate Bayesian inference in deep learning, ensembles are generally reported to achieve the best results in terms of predictive performance and reliability of the produced uncertainty estimates \citep{Ovadia2019, Gustafsson2020, Ashukha2020}.

\section{Data}
\label{sec:data}
\subsection{ALS structural forest variables}
\label{sec:als_data}
\begin{figure}
	\centering
	\includegraphics[width=\linewidth]{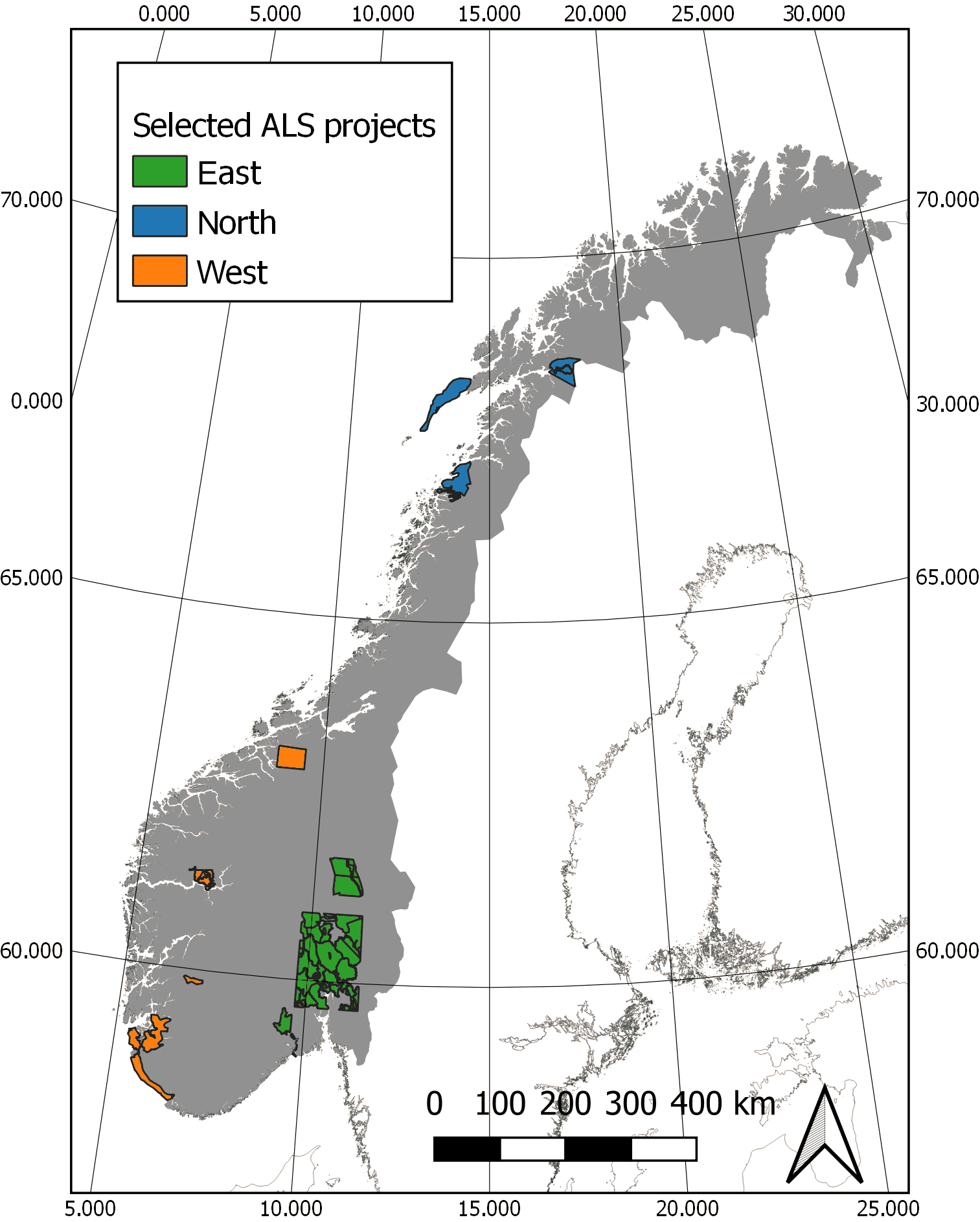}
	\caption{Overview of the selected ALS project used in this study for training, validation, and testing of the developed methods. The ALS projects are colored based on their geographical grouping.}
	\label{fig:als_regions}
\end{figure}

In this study, the target variables are forest structural variables normally extracted from ALS data. In Norway, the corresponding ALS data are available as part of a national program aimed at the production of a high resolution digital terrain model (DTM) of the entire country. Amongst the many available ALS projects, we selected a sample of 41 projects covering a range of latitude (58$^{\circ}$N – 69$^{\circ}$N) , longitude (5$^{\circ}$E-18$^{\circ}$E), and the corresponding diversity in landscapes and forest types in Norway. Our selected areas are shown in Fig.~\ref{fig:als_regions} and are covered by the following main forest types: 
\begin{itemize}
    \item East: productive boreal forest in the south eastern part, characterized by mild slopes and continental climate.
    \item North: predominantly deciduous areas with a large portion of low-productivity mountain birch (\emph{Betula nana}) in the north. These forests represent the transition from subpolar oceanic vegetation towards the tundra and sub-artic climates. 
    \item West: deciduous forest alternating with patches of coniferous forests and un-productive forests in the coastal areas. These areas are characterized by a milder oceanic climate that, due to the steep slopes (i.e., fjords), can transition to tundra and alpine vegetation within short distances.
\end{itemize}

The selected samples of ALS data were collected between 2015 and 2018, with $\sim$80\% being from the 2016 - 2017 period. For $\sim$35\% of the area, the ALS data were collected under leaf-off conditions (Oct.-Dec.). The point density in the selected ALS projects was 2 pts./m\textsuperscript{2} for 51\% of the area, 5 pts./m\textsuperscript{2} for a further 48\%, and 10 pts./m\textsuperscript{2} for the remaining 1\%. 
Overall, the sample covers a broad range of climatic, vegetation, terrain, and ALS data characteristics, thus providing a suitable training set for deep learning models that shall generalize across a range of conditions. 

\begin{figure}
	\centering
	\includegraphics[width=0.49\linewidth]{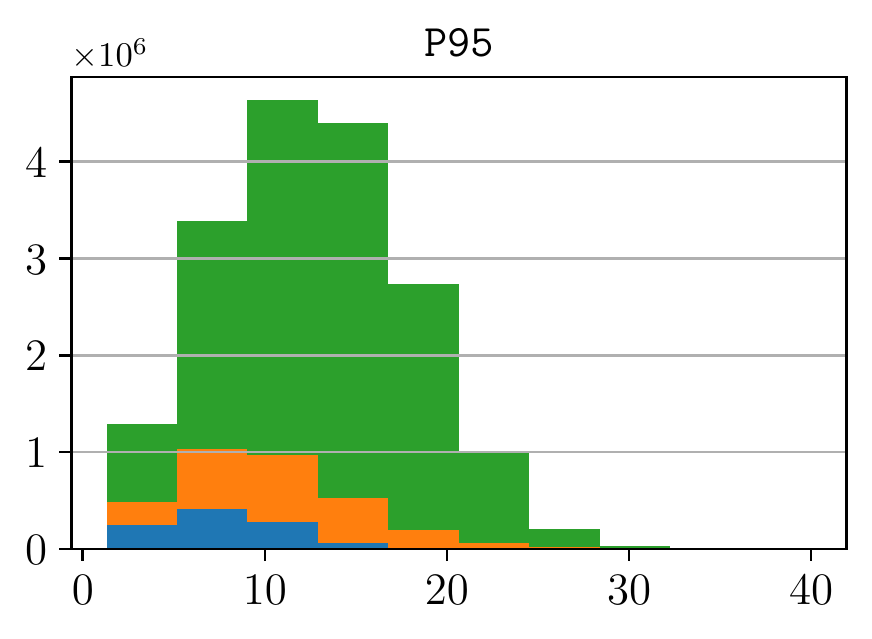}
	\includegraphics[width=0.49\linewidth]{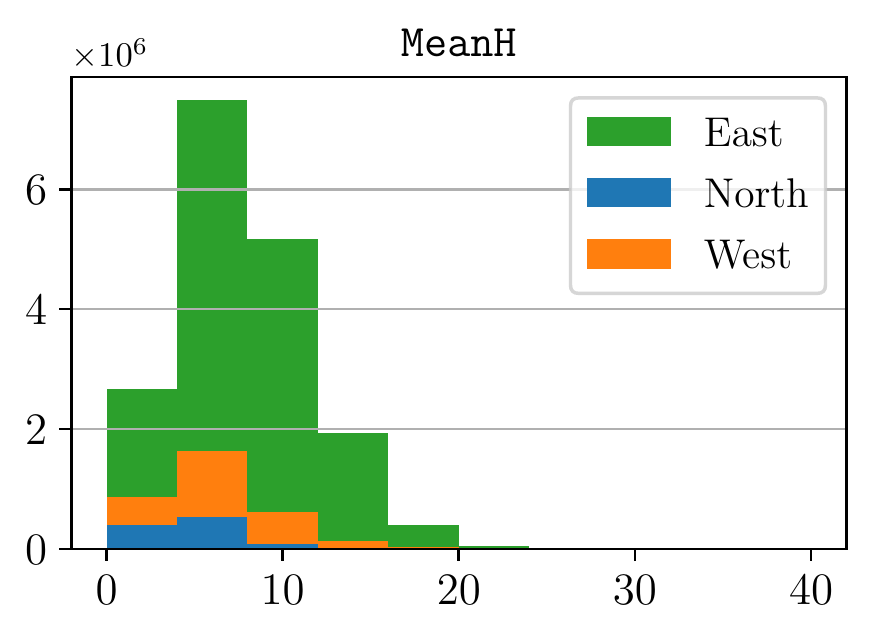}\\
	\vspace{-0.2em}
	\includegraphics[width=0.49\linewidth]{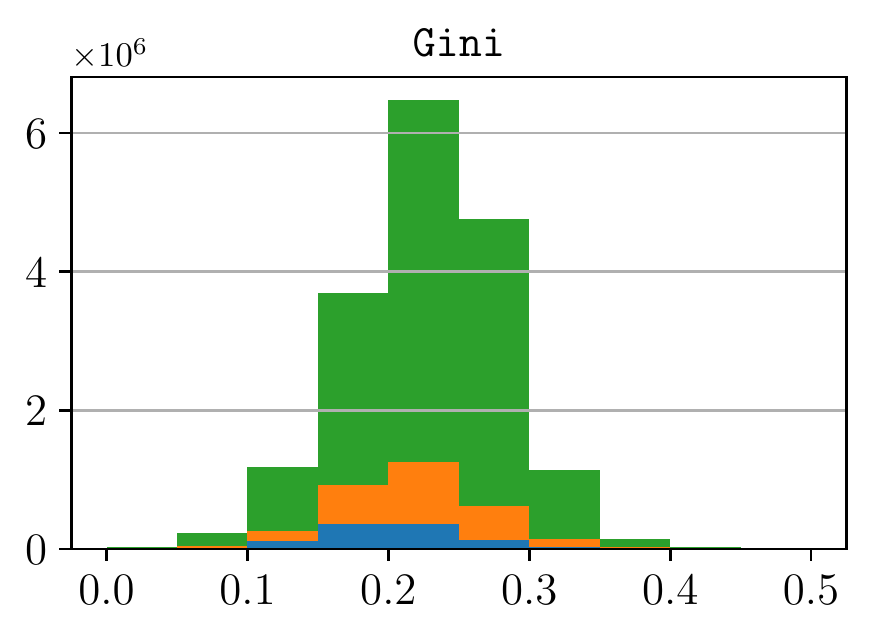}
	\includegraphics[width=0.49\linewidth]{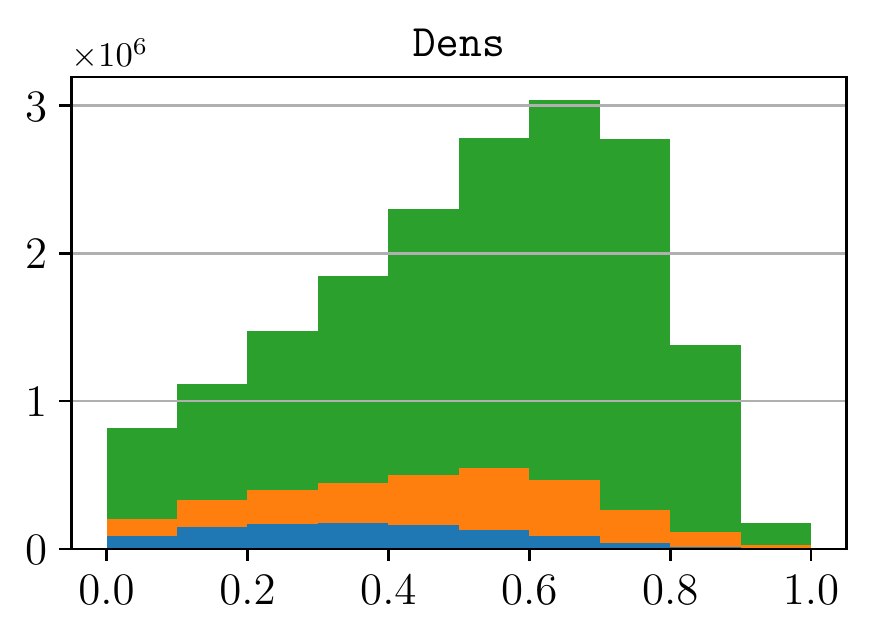}\\
	\vspace{-0.2em}
	\includegraphics[width=0.49\linewidth]{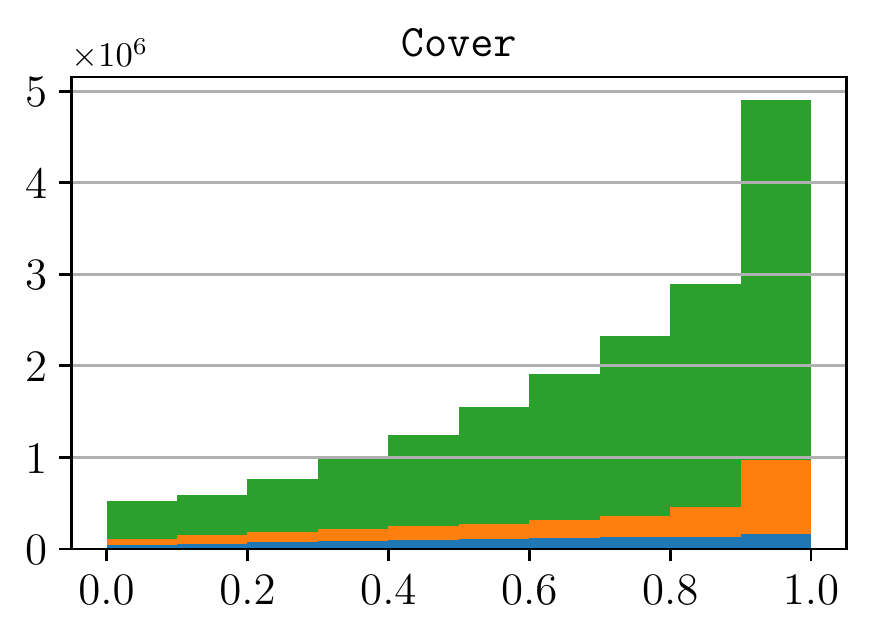}
	\caption{Histograms of the five ALS-derived structural forest variables. The coloring indicates the different geographical areas, and bars are stacked such that the overall bar height denotes the overall number of data points (i.e., pixels) for a given bin.}
	\label{fig:histograms}
\end{figure}

\begin{figure}
	\centering
	\includegraphics[width=\linewidth]{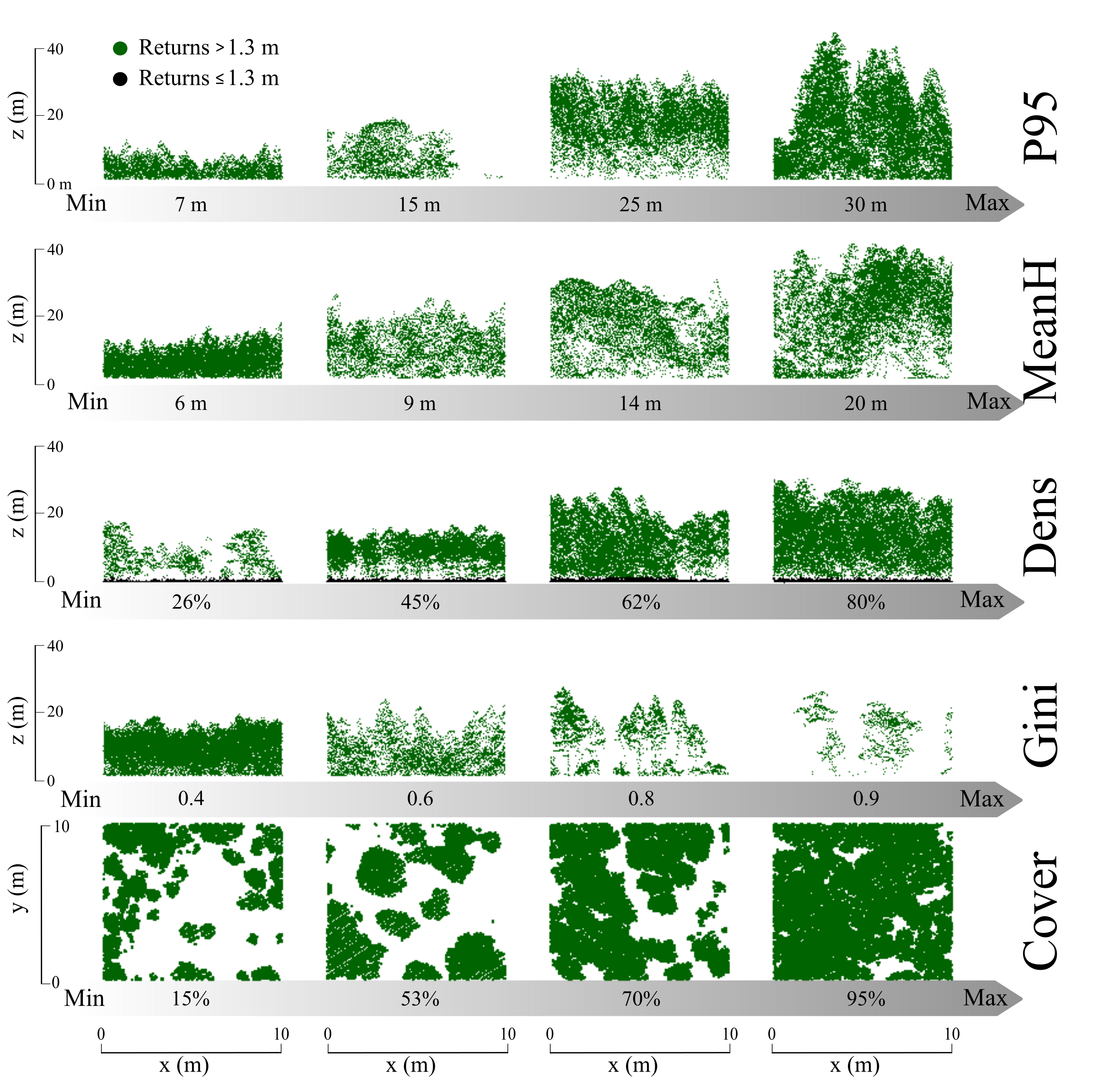}
	\caption{Visual representation of the variability in forest structures captured in the range of the studied ALS forest structural variables, including the 95\textsuperscript{th} percentile (\texttt{P95}; m), mean height (\texttt{MeanH}; m), vegetation density (\texttt{Dens}), Gini coefficient of the Dz values (\texttt{Gini}, \%), and forest cover (\texttt{Cover}; \%). For \texttt{P95}, \texttt{MeanH}, \texttt{Dens}, and \texttt{Gini} the scatter plots represent a side view of the normalized point clouds corresponding to Sentinel-2 pixels, while for the Cover the plots represent top-down view of plots representing the canopy a (in green). }
	\label{fig:als_vars}
\end{figure}

The height values $z$ (in meters above sea level) from the raw ALS point clouds were normalized  to $Dz$ (in meters above ground) by subtracting the ground elevation from each point’s $z$ value. Furthermore, all points with $Dz>$\,1.3\,m were classified as vegetation points. We extracted a selection of commonly used ALS variables to describe forest structure and ultimately ecosystem characteristics \citep{VALBUENA2020}: height, density/cover, complexity, and habitat area. These variables were selected with the aim to minimize the number of variables while maintaining complementary information on the vertical and horizontal distribution of the forest canopy. While these variables do not provide direct quantitative biophysical properties used in traditional forest management inventories (e.g. above ground biomass), they remain useful proxies for characterizing for forest monitoring purposes \citep{Senf2022, Hansen2013}. The variables were computed from the normalized point clouds (i.e., m above ground) on a 10~m raster, which was then bi-linearly resampled to match the Sentinel-2 pixels. The following variables were computed for the entire area where ALS was available. We show corresponding histograms in Fig.~\ref{fig:histograms} and an illustration of the variables in Fig.~\ref{fig:als_vars}.
\begin{itemize}
    \item \texttt{P95} (m above ground): the 95\textsuperscript{th} percentile of the $Dz$ values of all vegetation points. This variable is close to the maximum height but it removes potential noise from spurious high ALS returns (e.g., from birds or power lines). The use of the height percentiles dates back to the some of the first area-based forest inventories \citep{Naesset2002}. Among them, the 95\textsuperscript{th} percentile is useful as a measure of canopy top height, which is known to be correlated with the developmental stage of the forest and hence its biomass stock. Furthermore, canopy height diversity has been shown to be useful to monitor forest disturbance regimes \citep{Senf2020}.
    \item \texttt{MeanH} (m above ground): mean of the $Dz$ values of vegetation points. This variable is of interest because it not only captures the height of a forest but also the vertical distribution of the plant material within the canopy. We note that, thanks to its ability to encompass these two levels of information, \texttt{MeanH} is at present the only predictor variable used to produce country-wide maps of forest biomass in Norway \citep{Astrup2019}.
    \item \texttt{Dens} (\%): proportion of vegetation points in the entire set of LiDAR returns. This variable describes the density of the forest canopy and, along with the height percentiles, has long been used in area-based ALS forest inventories. Its is complementary to the canopy height, as it provides information about the vertical distribution of the plant material through the canopy \citep{Naesset2002}. 
    \item \texttt{Gini} (index): The Gini coefficient is equal to half of the relative mean difference in $Dz$ values among all the vegetation returns, and was calculated using the function implemented in the \emph{leafR} package \citep{Almeida2020}. The Gini coefficient is a measure of the inequality among the members of a data distribution, and it has been used as a proxy for tree size variation \citep{Knox1989} and to map differences in forest structures and management regimes \citep{Valbuena2017, Adnan2019}. While typically the Gini coefficient has been calculated using single-tree data, a recent work by \citep{Adnan2021} demonstrated the usefulness of the Gini coefficient calculated from the ALS $Dz$ values, showing that it could reliably describe the structural heterogeneity of the forest.
    \item \texttt{Cover} (\%): Forest cover in terms of the proportion of projected canopy area relative to the entire area of a pixel (100 m\textsuperscript{2}). Cover was computed by projecting the vegetation points onto a $(x,y)$ plane and converting them to a binary occupancy grid with 1 m resolution. The forest cover was then derived as the percentage of pixels occupied by forest. While somewhat correlated to \texttt{Dens}, cover remains a fully 2-dimensional variable, describing the horizontal vegetation cover, rather than the density of points in the vertical canopy profile. Time series of forest cover maps have been widely used in remote sensing as a measure of the canopy openness, and to assess land use changes with particular interest on forest cover losses \citep{Hansen2013} and forest disturbance \citep{Senf2018}. 
\end{itemize}

We divide each ALS project geographically into horizontal stripes of width 900 pixels (9.0 km). Within each stripe we assign the northernmost 5.4 km to the training set, the next 1.8 km to the validation set and the southernmost 1.8 km to the test set. In this way, the different regions and modalities in each ALS area are evenly distributed between the three sets. Overall, the data set consists of 105,022,419 pixels (10,502 km\textsuperscript{2}) of ALS reference data, divided into 64,487,551 training pixels (6,449 km\textsuperscript{2}), 20,784,407 validation pixels (2,078 km\textsuperscript{2}) and 19,750,461 test pixels (1,975 km\textsuperscript{2}), following the ``60-20-20" ratio often used in machine learning studies.

\subsection{Sentinel satellite imagery}
Sentinel-1 and Sentinel-2 are satellite missions that belong to the European Union's Copernicus Earth observation programme \citep{cop_ref}, providing high-resolution SAR and optical imagery of land and coastal water areas between 56° South and 84° North. Each mission consists of a ground segment and a constellation of two satellites in sun-synchronous low earth orbits, phased 180° from each other. Jointly, both satellites in each mission achieve high revisit frequencies of $>$1 visit every five days, depending on latitude. The Sentinel-2 satellites are equipped with multispectral instruments that detect light in 13 spectral bands ranging from visible blue to short-wave infrared, with band-dependent spatial resolutions between 10 and 60 m. This spectral profile gives rise to a multitude of applications, including forest and vegetation monitoring as well as the inference of vegetation parameters such as leaf area index or carbon stock. For this work, we have collected Sentinel-2 Bottom-of-Atmosphere (BOA) reflectance images \citep[level 2A,][]{Main2017} that have already been atmospherically corrected. For each ALS acquisition area, we select largely cloud-free images captured between May and October of the respective year, such that the area of interest is fully covered. Because BOA reflectance images still contain small amounts of atmospheric variation, we collect between two and seven images for each ground point (depending on cloud conditions), such that our model learns to cover a range of conditions. \\

The Sentinel-1 satellites are equipped with a C-band SAR sensor that actively monitors the surface with radiation of about 6 cm wavelength \citep{s1_url}. Sentinel-1 is invariant against meteorological factors like clouds, and also against illumination conditions.
Although C-Band SAR does not enter deep into the canopy, it may superficially penetrate it. Thus, we explore this as a complementary signal to the optical image.
For every collected optical image, we query a Sentinel-1 image that was acquired within ten days of the optical image and that covers the same geographic area. As preprocessing, we carry out orbit correction and terrain correction, where for the latter we rely on the Copernicus digital elevation model with a resolution of one arc second. We intentionally keep the preprocessing simple because we expect our model to automatically infer the optimal transformations given the task at hand. For all preprocessing steps, we use SNAP~\citep{snap_url} with the Sentinel-1 Toolbox~\citep{S1toolbox_url} as provided by ESA.

\section{Method}
\label{sec:method}

\subsection{Forest structure model architecture}
\begin{figure*}
	\centering
	\includegraphics[width=\linewidth]{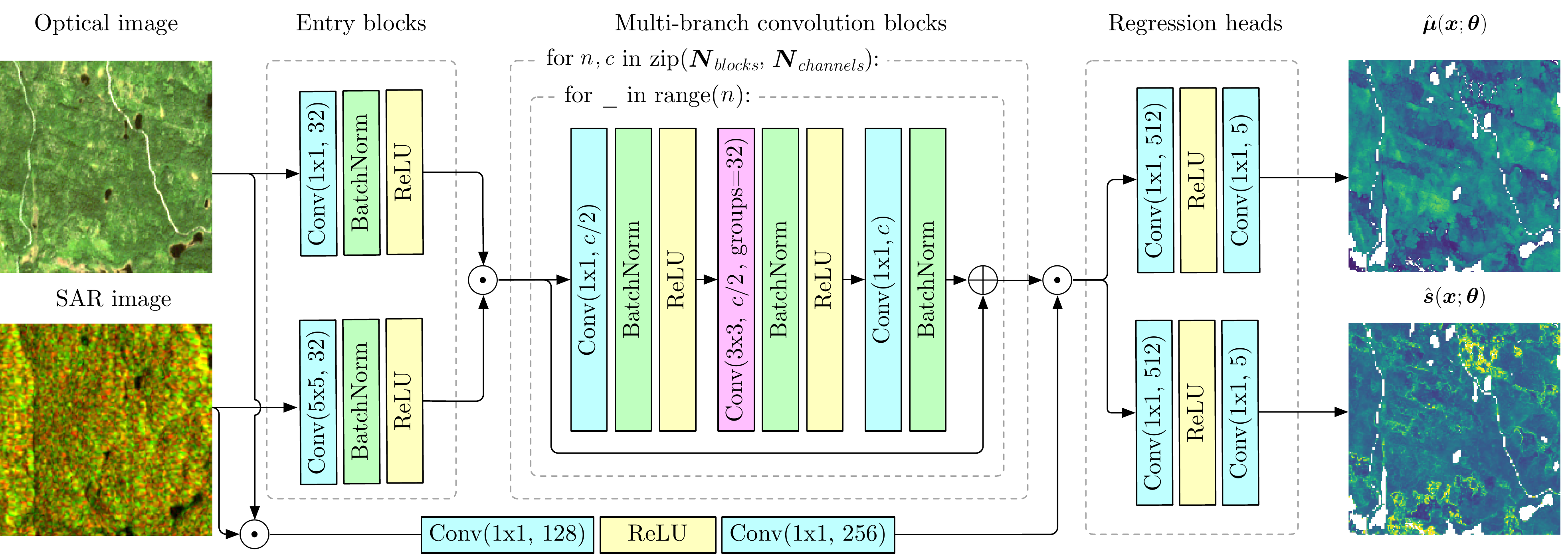}
	\caption{\setlength{\fboxsep}{1pt}Proposed model architecture. Convolutional layers are shown in \colorbox{cyan}{cyan}, with the kernel size and number of output channels in brackets. Batch normalization layers are in \colorbox{lime}{lime}, ReLU activations in \colorbox{yellow}{yellow} and  grouped convolutions in \colorbox{purple}{purple}, with an additional argument for the number of groups. $\oplus$ and $\odot$ denote element-wise addition, respectively concatenation along channel dimension, branching arrows indicate sharing (``copying") of the respective tensor. In our experiments we set $\bm{N}_\text{blocks} = [2, 3, 5, 3]$ and $\bm{N}_\text{channels} = [256, 512, 1024, 2048]$.}
	\label{fig:architecture}
\end{figure*}

We utilize a multi-branch convolutional network (CNN) to transform co-registered optical and SAR images into maps of forest structure variables, and of their associated uncertainties. A CNN gradually transforms its inputs into outputs with a series of simple transformations (layers). One can think of the early layers as a pre-processing of the two input modalities, the middle layers as a feature extractor and the last layers as a regression, but there is no clear distinction between those parts.
At the heart of all of them is the (discrete) convolution operator, i.e., a linear filtering of a $c$-channel input image with a $k \times k \times c$ kernel of weights. In each convolutional layer, multiple convolutions with different filter weights are applied, so the output is again an ``image" with as many channels as there were filters.
The filter weights are the trainable parameters of the model. Convolutional layers are interleaved with some element-wise, non-linear activation functions. The series of transformations gradually abstracts the bare image pixels into a sequence of feature maps until, at the output level, they have become the desired forest structure maps.

Our model architecture is inspired by the design principles of ResNeXt \citep{Xie2017}, modified to address the specific challenges of forest structure mapping. As in a number of remote sensing works \cite[e.g.,][]{Rodriguez2018,Lang2019}, we retain the spatial dimensions of the input images throughout the network, i.e., we do not use any pooling operations or convolutions with stride. This is motivated by the observation that spatial feature aggregation tends to negatively impact performance on remote sensing images, as their spatial resolution is already limited and one cannot afford to lose further spatial detail.
Furthermore, we have added separate \emph{entry blocks} for the optical and SAR channels, allowing for domain-specific low-level processing.
Finally, we have added a \emph{global pixel shortcut} that re-injects the raw pixel values before the final regression layers, in order to better preserve high-frequency details that otherwise tend to get blurred (as information over an increasingly larger receptive field is mixed  through repeated convolutions). We visualize our model architecture in Fig.~\ref{fig:architecture} and in the following, the data flow in the model is described in more detail. 

\paragraph{Entry blocks}
Each entry block consists of a convolutional layer, followed by batch normalization \citep{Ioffe2015} and a rectified linear unit (ReLU) non-linearity, which truncates negative values to zero. The convolution in the optical entry block has kernel size 1, i.e., it only mixes the channels at every individual pixel, while a larger kernel size of five is used in the SAR entry block to allow for learned spatial smoothing of the more noisy SAR data. For all convolution operations in our model that have kernel size \textgreater1, we apply appropriate zero-padding to preserve the spatial dimensions of the input.

\paragraph{Multi-branch convolution blocks}
The outputs of both entry blocks are concatenated along their channel dimension and fed through $N_\text{stages}$ \emph{stages}, where each stage $i \in \{1\dots N_\text{stages}\}$ consists of $N_{\text{blocks},i}$ ResNeXt \emph{blocks}. Each ResNeXt block in turn is made up by a $1 \times 1$ convolution layer, a grouped convolution layer \citep{Krizhevsky2012, Xie2017} with $N_\text{groups}$ groups and kernel size three, and another $1 \times 1$ convolution layer. The former two are each followed by a batch normalization and ReLU, whereas the latter is followed only by batch normalization. Following the ResNet principle, a skip connection bypasses each block, such that the block learns an additive residual update to its input, making it possible to training much deeper networks \citep{He2015}. Due to the grouped convolution, each block effectively implements a multi-branch computational graph, where each branch can be interpreted as a lower-dimensional feature embedding, and where all branches are eventually combined by summation. This layout has been shown to offer superior predictive performance \citep{Xie2017}, while being less complex than the classical ResNet design \citep{He2015}.

\paragraph{Regression heads}
Finally, from the resulting feature map, two parallel \emph{heads} regress the parameters of the likelihood function for each structural variable and pixel. We assume a Gaussian likelihood, therefore for every pixel five means and log-variances are regressed. The variance of each predictive distribution will be trained to resemble aleatoric uncertainty, and is output in its logarithmic form for numerical stability (see Sec.~\ref{sec:loss}).
Each head consists of two convolution layers with kernel size one and an intermediate ReLU operation, gradually reducing the number of channels in the representation from 2304 down to five. In order to constrain the predicted mean values to their valid ranges (see Sec.~\ref{sec:data}), we apply a final $\exp(\cdot)$ activation to the \texttt{P95} and \texttt{MeanH} predictions, as well as a sigmoid activation $\sigma(x) = 1 \mathbin{/} (1 + e^{-x})$ to the \texttt{Dens}, \texttt{Gini} and \texttt{Cover} means. 

\subsection{Model training}
\label{sec:model_training}
As commonly done in deep learning, we iteratively learn the model parameters with stochastic gradient descend, starting from a random initialization. In each iteration, we randomly sample a batch of $B=64$ reference data patches of size $15 \times 15$ pixels, where a patch is only considered for training if the center pixel is \emph{forested}. We consider a pixel forested if and only if it contains vegetation points (points with $Dz>$\,1.3\,m, see Sec.~\ref{sec:als_data}) and also is considered forested based on NIBIO's Norway-wide timber volume map \citep{Astrup2019}. We use the latter as an additional precautionary measure to avoid unnecessary noise from non-forested areas, as  we are interested in learning forest characteristics only.
For every reference data patch, we randomly pick an optical image from the correct year and two SAR images (one ascending and one descending orbit) with acquisition dates near the one of the optical image. Using SAR with both ascending and descending orbits is expected to add robustness against terrain-induced geometric distortions such as shadowing, foreshortening and layover \citep{Small1995, Carrasco1997}.
In total the model input is composed of 12 optical bands forming a tensor of size $B\!\times\!12\!\times\!15\!\times\!15$, and four SAR bands (two polarizations $\times$ two orbital directions), forming a corresponding $B\!\times\!4\!\times\!15\!\times\!15$ tensor. The model output are two tensors of shape $B\!\times\!5\!\times\!15\!\times\!15$ each, one for the means of the five structural forest variables and one for their variances. An appropriate \emph{loss function} measures the deviation between the model output and the ALS reference data, see below. Note that we only calculate this loss for pixels that are forested according to the above definition. Then, the gradients of all model parameters $\bm{\theta}$ w.r.t.\ that loss (plus some regularization term) are computed with back-propagation \citep{Werbos1982} and used to update the parameters in the direction of steepest descent. We use the Adam \citep{Kingma2014} variant of stochastic gradient descent (SGD), which adaptively scales the magnitude of the parameter updates based on the statistics of previous updates to speed up convergence. During training, we periodically evaluate the prediction error of the model (i.e., the current set of parameters) on a held-out validation set and keep the configuration $\bm{\theta}^*$ with the lowest error as the final model.

\subsection{Loss function}
\label{sec:loss}
The loss function, which is optimized during training, measures the quality of a set of network parameters $\bm{\theta}$ w.r.t. the training data $\mathcal{D} = \{ (\bm{x}_i, \bm{y}_i) \}_{i=1}^N$, under some regularizing prior assumptions. We use a standard loss function $\mathcal{L}(\mathcal{D}; \bm{\theta})$ whose minimization corresponds to maximizing the posterior probability of the parameters given the training data. As it is commonly done in machine learning (see e.g. \citet{Goodfellow2016}), we assume a zero-mean isotropic Gaussian prior over the network parameters (corresponding to $\mathcal{L}_2$ regularization) and a Gaussian likelihood function with mean $\hat{\bm{\mu}}_i := \hat{\bm{\mu}}(\bm{x}_i; \bm{\theta}) \in \mathbb{R}^5$ and diagonal covariance matrix with logarithmic elements $\hat{\bm{s}}_i := \hat{\bm{s}}(\bm{x}_i; \bm{\theta}) \in \mathbb{R}^5$:
\begin{align}
&\mathcal{L}(\mathcal{D}; \bm{\theta}) = \lambda \left\|\bm{\theta}\right\|_2^2 + \sum_{i,j} \big[ \hat{s}_{ij} + \exp(-\hat{s}_{ij}) (\hat{\mu}_{ij} - y_{ij})^2 \big]
\label{eq:loss}
\end{align}
Here, $i \in \{1\dots N\}$ indexes the data point while $j \in \{1\dots 5\}$ indexes one of the five structure variables we are aiming to predict. The hyperparameter $\lambda$ is inversely proportional to the variance of the parameter prior. Note that by explicitly predicting aleatoric uncertainty, in the form of log-variances $\hat{\bm{s}}(\bm{x}_i; \bm{\theta})$ of the output variables, the model learns to reduce the influence of data points on the loss that it deems particularly noisy, which in turn improves model performance \citep{Kendall2017}. Learning log-variances constrains the variances to positive values and prevents a potential division by zero in the loss function. 
The derivation and some further explanation of the loss function Eq.~\ref{eq:loss} is provided in \ref{ap:loss}.

\subsection{Acquiring uncertainty estimates}
Let $\hat{\bm{\mu}}(\bm{x}_*;\bm{\theta})$ and $ \hat{\bm{\sigma}}^2(\bm{x}_*;\bm{\theta})$ be the mean and variance of the predicted distribution that the network with parameters $\bm{\theta}$ outputs when shown the test image $\bm{x}_*$. Furthermore, let $p(\bm{\theta} \mid \mathcal{D})$ denote the posterior distribution over the network parameters, given the prior and the likelihood function defined in Sec.~\ref{sec:loss}. 

The exact predictive distribution $p(\bm{y}_* \mid \bm{x}_*,\mathcal{D})$ of our model is intractable to compute, mainly due to the extremely high dimensionality of the parameter space. 
However, it is possible to \emph{sample} from an approximate $p(\bm{\theta} \mid \mathcal{D})$ by training an ensemble of multiple neural networks from the same data (but with different random initializations and random batches for SGD). With $M$ different networks this gives rise to a Monte Carlo approximation of the posterior,
\begin{align}
p(\bm{y}_* \mid \bm{x}_*,\mathcal{D}) \approx \frac{1}{M} \sum_{k=1}^M \mathcal{N}\big(\bm{y}_* \mid \hat{\bm{\mu}}_{*,k}, \text{diag}(\hat{\bm{\sigma}}^2_{*,k})\big),
\label{eq:monte_carlo}
\end{align}
where $\hat{\bm{\mu}}_{*,k} := \hat{\bm{\mu}}(\bm{x}_{*};\bm{\theta}_k)$ and $\hat{\bm{\sigma}}^2_{*,k} := \hat{\bm{\sigma}}^2(\bm{x}_{*};\bm{\theta}_k)$ are the mean and variance predicted by the $k$-th network. \\

In the approximate predictive distribution derived in Eq.~\ref{eq:monte_carlo}, epistemic (i.e., model) uncertainty is captured by sampling multiple models $\bm{\theta}_k$, which will lead to widely scattered predictions and thus to high uncertainty in regions of the input space that are not sufficiently backed by training data. Aleatoric uncertainty, on the other hand, is learned from data and thus predicted as the variance of the likelihood function for each output variable at each image pixel.
Note that for conceptual and computational simplicity, the likelihood is limited to have a diagonal covariance matrix.
While this restricts the network's flexibility to express correlations in aleatoric uncertainty, we empirically find this modeling decision to be sufficient for the task at hand.
Finally, based on Eq.~\ref{eq:monte_carlo}, the tower rule and the law of total variance deliver approximations for the total mean and variance of the predictive distribution:
\begin{align}
    \mathbb{E}[\bm{y}_* \mid \bm{x}_*,\mathcal{D}] &\approx \frac{1}{M} \sum_{k=1}^M \hat{\bm{\mu}}_{*,k} =: \bar{\bm{\mu}}_* \label{eq:pred_mean}\\
    \text{Var}[\bm{y}_* \mid \bm{x}_*,\mathcal{D}] &\approx \frac{1}{M} \sum_{k=1}^M [ \hat{\bm{\sigma}}^2_{*,k} + (\hat{\bm{\mu}}_{*,k} - \bar{\bm{\mu}}_*)^2 ] =: \bar{\bm{\sigma}}^2_* \label{eq:pred_var}
\end{align}
In Sec.~\ref{sec:uncertainty}, we will empirically study the quality of the uncertainty estimate (\ref{eq:pred_var}), and demonstrate how it can be used to identify and filter out unreliable predictions.

\subsection{Implementation details}
We have implemented our models in PyTorch \citep{Paszke2017}. We trained $M = 5$ models with batch size $B = 64$ and a base learning rate $\alpha = 10^{-4}$. The learning rate is automatically reduced by a factor of 0.1 when the validation loss has not improved for 15 consecutive epochs. We apply weight decay to control the strength of the unit Gaussian prior, with an empirically chosen magnitude of $10^{-3}$ that is inversely proportional to the hyperparameter $\lambda$ from Eq.~\ref{eq:loss}. We chose $\beta_1 = 0.9$, $\beta_2 = 0.999$ and $\epsilon = 10^{-8}$ as hyper-parameters for the Adam optimizer. Each neural network was trained on a single Nvidia RTX2080Ti GPU for $\sim$14 days.

\section{Experimental Results and Discussion}
\label{sec:results_and_discussion}
\begin{figure*}[th!]
	\centering
	\includegraphics[width=\linewidth]{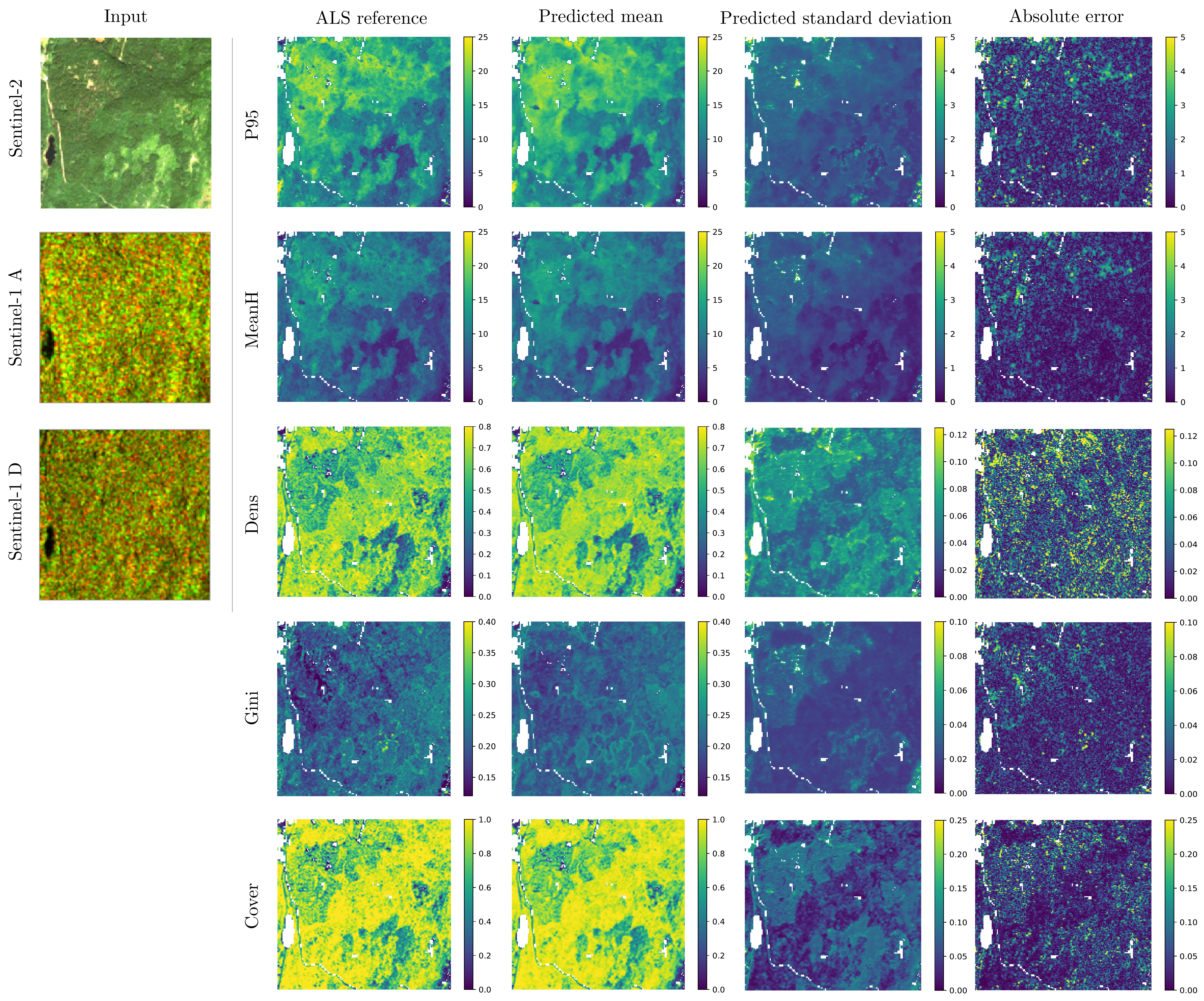}
	\caption{Example result from the test set of the \emph{East} area ($180 \times 180$ pixels, 324 ha). \emph{Left tile}: The model input is composed of a Sentinel-2 optical image (12 bands), and two preprocessed Sentinel-1 SAR images taken from an ascending and descending orbit direction, respectively. For visualization, the amplitudes of the VH and VV polarizations have been assigned to the red and green channels of the RGB image. \emph{Right tile}: ALS reference data, the predicted mean and standard variation as well as the absolute prediction error $|\bar{\mu}_{ij} - y_{ij}|$ for each of the five structural variables. The figure shows high correspondence between reference data and the predicted mean, indicating that our model is able to accurately regress the structural variables. There is also a clear (albeit not as crisp) correlation between predicted standard variation and absolute error, indicating that erroneous predictions are assigned higher uncertainty. We analyze the latter in more detail in Sec.~\ref{sec:uncertainty}.}
	\label{fig:prediction_sample}
\end{figure*}

During evaluation, we slide a window of size $15 \times 15$ pixels over the test regions of our data. We use a stride of nine and only retain the innermost $11 \times 11$ pixels for every window location, thus allowing for a sufficiently large spatial input context for all predictions. Fig.~\ref{fig:prediction_sample} depicts the model output for a diverse 180 $\times$ 180 pixel example region sampled from the \emph{East} area of the test set, along with the ALS reference data and the absolute error between the predicted mean and the reference data. Qualitatively, a high correspondence between the predicted mean and the reference data can be observed, indicating that the model has successfully learned to predict the respective variables for a very diverse sample region. Further, a correlation between predicted uncertainty and absolute error can be observed, although the latter is more ``grainy", an attribute that we believe is transferred from the reference data, that can vary strongly between adjacent pixels (this is not that visible as in the error maps, because the overall value range is larger). The correlation between those quantities is not as visually obvious as the previous one, which calls for a more extensive investigation into the predicted uncertainties. Similar to Fig.~\ref{fig:prediction_sample}, we have added qualitative examples from the \emph{North} and \emph{West} areas in Figs.~\ref{fig:prediction_sample_north}
and \ref{fig:prediction_sample_west} in the appendix.

In the following subsections, among other experiments, we will quantitatively validate the observations made above (good reconstruction of the true structural variables as well as uncertainty estimates being representative of the error) for the entire test set.

\subsection{Evaluation of forest variables prediction}
\label{sec:eval}

For each variable, indexed by $j \in \{1\dots5\}$, we report the Mean Absolute Error (MAE), Root Mean Square Error (RMSE) and the Mean Bias Error (MBE) defined as follows:
\begin{align}
\text{MAE} = \frac{1}{N_{test}}\sum_{i=1}^{N_{test}} |\bar{\mu}_{ij} - y_{ij}| \\
\text{RMSE} = \sqrt{\frac{1}{N_{test}}\sum_{i=1}^{N_{test}} (\bar{\mu}_{ij} - y_{ij})^2} \\
\text{MBE} = \frac{1}{N_{test}}\sum_{i=1}^{N_{test}} \bar{\mu}_{ij} - y_{ij}
\end{align}
To represent the predictive distribution for each pixel $i$ and variable $j$ as a point estimate, we use its approximate mean $\bar{\mu}_{ij}$ in accordance with Eq.~\ref{eq:pred_mean}. The MAE and RMSE metrics measure the average deviation between model prediction and reference data (with RMSE giving higher penalty to large deviations), while the MBE serves to identify systematic biases in the predictions. In addition to the above metrics, we calculate their \emph{normalized} counterparts MAE\%, RMSE\% and MBE\% by dividing by the corresponding mean values over the training data. 

Tab.~\ref{tab:results_all} shows the results of the test set evaluation of the final model. Overall (subtable a), we achieve low mean absolute errors for all predicted variables, e.g., 1.65 m for the 95th height percentile (\texttt{P95}). Assessing the relative errors, the experiments show that the developed model produces consistently accurate predictions with an MAE\% always smaller than 15\%. The variables ranking in order of increasing relative error (MAE\%) are forest cover (\texttt{Cover}, 11.2\%), vegetation density (\texttt{Dens}, 11.8\%), Gini index (\texttt{Gini}, 12.4\%), the 95th height percentile (\texttt{P95}, 12.9\%) and lastly the mean height (\texttt{MeanH}, 14.4\%). 
When evaluating the individual geographic regions shown in Fig.~\ref{fig:als_regions}, we observe comparable values for the eastern region (subtable b), which can be explained by this region making up for the majority of reference data points. Interestingly, the model performs even better in the northern test region in terms of mean absolute error of \texttt{P95} and \texttt{MeanH}, however the error for the remaining structural variables is slightly bigger.
For the western test region, evaluation errors are consistently higher, yielding e.g. an MAE in \texttt{P95} of 1.85 meters and an MAE\% of 14.4\%. Possible reasons for the slightly lower performance in the western region are the scarcity of cloud-free data due to oceanic climate, and the increased topographic complexity. The fjord landscape in these western areas is characterized by very steep and narrow valleys causing both topographic shading and large variations in sun-target-sensor geometry. All of these factors have a negative impact on the quality of the satellite observations and thus the predictions obtainable in such areas. In addition, the forests in these areas are more diverse in terms of species compositions and structurally complex compared to the other areas.
It is noteworthy that, for all of the tested regions, the MBE\% is $<$0.7\%, indicating the absence of any systematic model bias, regardless of the predicted variable and geographic region. Our model can therefore be considered an unbiased estimator for forest structure, a significant benefit for downstream tasks such as decision making based on the resulting maps.

\begin{figure*}
	\centering
	\includegraphics[width=\linewidth]{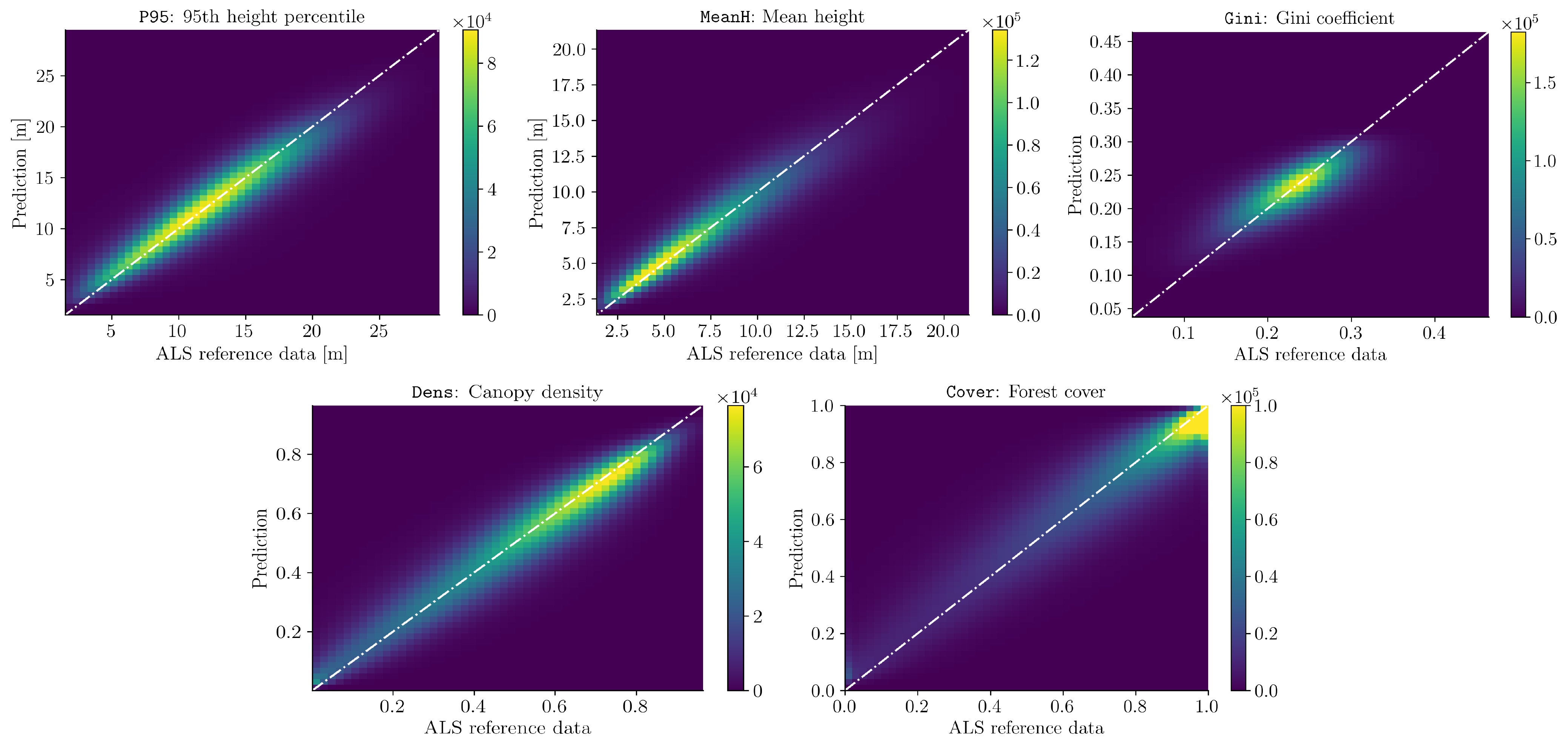}
	\caption{Confusion plots for the predicted structure variables. In each plot, the $x$ axis denotes the ALS reference data value and the $y$ axis denotes the mean of the predicted likelihood distribution. The number of data points that fall into each 2D bin is indicated by the color. The plots demonstrate that most predictions are located very close to the identity line, where prediction and ALS reference data agree.}
	\label{fig:confusion}
\end{figure*}

For further analysis of our model, we provide confusion plots for all variables in Fig.~\ref{fig:confusion}. For the plots, all test samples are binned with regard to both their true, ALS-derived reference value ($x$ axis) and their predicted mean ($y$ axis), the color of a bin indicating how many samples fall into the respective bin. Higher densities along the identity line imply higher accuracy of the model, and indeed we can observe very high agreement between prediction and reference data, for all variables.

\begin{figure*}
	\centering
	\includegraphics[width=\linewidth]{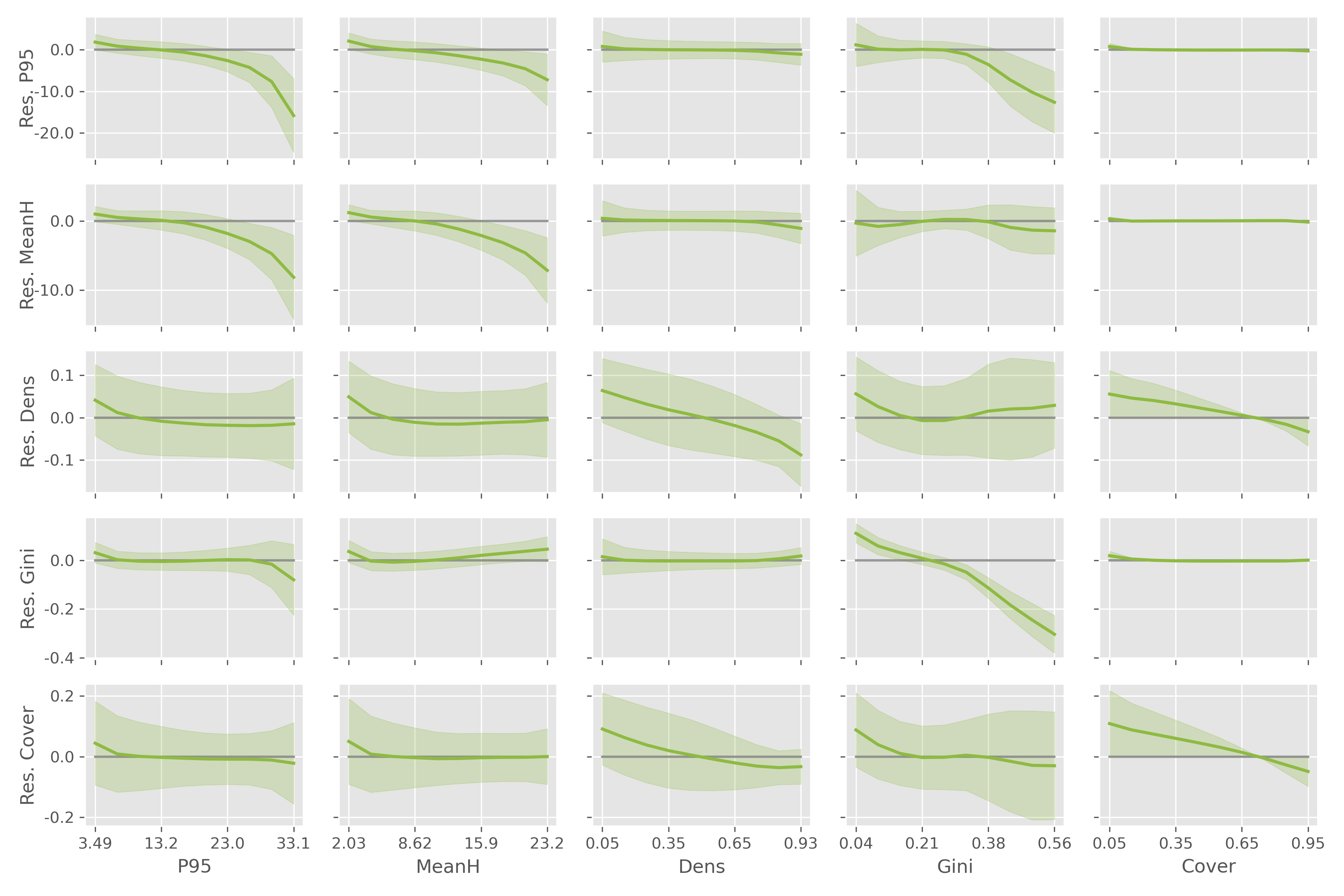}
	\caption{Residual structure analysis for all variables. Columns indicate the variable we bin according to (\emph{query}), and rows report the residuals of each variable in the respective bins (\emph{target}). Additionally, in shaded we show the respective standard deviation of the residuals.}
	\label{fig:res_analysis}
\end{figure*}

In addition, we conduct a residual structure analysis of each variable with respect to all others, i.e., we investigate how the residuals in one variable are distributed across other variables' value ranges.
Note that to get more stable estimates, we remove data points exceeding the 99th percentile of the respective variable before binning. For this analysis, we used $b=10$ bins.
In Fig.~\ref{fig:res_analysis}, the column indicates the \emph{query} variable, whose ground truth values (clipped to the 99th percentile for stability) are discretized into $b$ bins and define the ordering along the $x$-axis. The row indicates the \emph{target} variable whose residual distribution is plotted (so, for instance, the fourth graph in the top row are the \texttt{P95} residuals ordered by the corresponding pixels' \texttt{Gini} coefficient). The solid green line denotes the mean residual, the shaded area the residuals' standard deviation. Mean residuals close to zero (grey line) mean that the estimation of the target is unbiased at the given value of the query.
We observe that our predictions generally have low bias across a large portion of the range, which is consistent with our findings in Sec.~\ref{sec:eval}. A notable exceptions is the under-estimation of the height-related variables \texttt{P95} and \texttt{MeanH} on high trees -- a well-known effect when retrieving vegetation height from satellite images~\cite{potapov2021mapping,Lang2019}.
A similarly pronounced under-estimation bias at the top of their own value range can be noticed for the canopy density (\texttt{Dens}) and for the Gini coefficient (\texttt{Gini}); moreover, high \texttt{Gini} values also tend to cause underestimation of \texttt{P95}.
In the opposite direction, we also observe some (weaker) over-estimation biases for low reference values. This concerns in particular the \texttt{Dens} and \texttt{Cover} estimates at low values of \texttt{Dens}, \texttt{Gini} and \texttt{Cover}.
Overall, the biases are moderate and exhibit the typical behavior of regressors fitted to minimize mean squared error. That is, they tend towards the data mean and, when the evidence is weak or ambiguous, over-estimate low values and under-estimate high ones.

\begin{table}
    \centering
    \caption{Test set evaluation results of the proposed model for all ALS areas (subtable a) as well as for the individual geographic regions shown in Fig.~\ref{fig:als_regions} (subtables b - d). \texttt{P95} and \texttt{MeanH} metrics are given in meters, while \texttt{Dens}, \texttt{Gini} and \texttt{Cover} are reported as fractions. All normalized metrics (postfixed with \%) are given as a fraction over the respective training set mean.}
    \begin{tabular}{@{}p{0.1cm} c|c|c|c|c|c@{}}
        & & \texttt{P95} & \texttt{MeanH} & \texttt{Dens} & \texttt{Gini} & \texttt{Cover} \\
        \hline
        \multirow{6}{*}{\rotatebox[origin=c]{90}{a) All regions}} & MAE & 1.648 & 1.127 & 0.061 & 0.028 & 0.078 \\
        & MAE\% & 0.129 & 0.144 & 0.118 & 0.123 & 0.112 \\
        & RMSE & 2.298 & 1.595 & 0.082 & 0.039 & 0.107 \\
        & RMSE\% & 0.179 & 0.204 & 0.158 & 0.170 & 0.154 \\
        & MBE & -0.086 & -0.040 & -0.003 & -0.001 & 0.001 \\
        & MBE\% & -0.007 & -0.005 & -0.006 & -0.004 & 0.002 \\
        \hline
        \hline
        \multirow{6}{*}{\rotatebox[origin=c]{90}{b) East}} & MAE & 1.631 & 1.127 & 0.060 & 0.027 & 0.075 \\
		& MAE\% & 0.127 & 0.144 & 0.114 & 0.121 & 0.107 \\
		& RMSE & 2.250 & 1.589 & 0.080 & 0.038 & 0.101 \\
		& RMSE\% & 0.176 & 0.203 & 0.153 & 0.166 & 0.145 \\
		& MBE & -0.090 & -0.040 & -0.004 & -0.001 & 0.000 \\
		& MBE\% & -0.007 & -0.005 & -0.007 & -0.005 & 0.000 \\
        \hline
        \hline
        \multirow{6}{*}{\rotatebox[origin=c]{90}{c) North}} & MAE & 1.451 & 0.904 & 0.071 & 0.031 & 0.106 \\
		& MAE\% & 0.113 & 0.115 & 0.136 & 0.136 & 0.152 \\
		& RMSE & 2.238 & 1.329 & 0.095 & 0.042 & 0.140 \\
		& RMSE\% & 0.175 & 0.170 & 0.182 & 0.186 & 0.200 \\
		& MBE & -0.068 & -0.038 & 0.003 & 0.001 & 0.009 \\
		& MBE\% & -0.005 & -0.005 & 0.005 & 0.005 & 0.013 \\
        \hline
        \hline
        \multirow{6}{*}{\rotatebox[origin=c]{90}{d) West}} & MAE & 1.845 & 1.224 & 0.069 & 0.031 & 0.091 \\
		& MAE\% & 0.144 & 0.156 & 0.132 & 0.136 & 0.131 \\
		& RMSE & 2.609 & 1.740 & 0.093 & 0.042 & 0.127 \\
		& RMSE\% & 0.204 & 0.222 & 0.178 & 0.184 & 0.182 \\
		& MBE & -0.069 & -0.041 & -0.001 & -0.000 & 0.003 \\
		& MBE\% & -0.005 & -0.005 & -0.002 & -0.001 & 0.005 \\
    \end{tabular}
	\label{tab:results_all}
\end{table}

\subsection{Uncertainty evaluation}
\label{sec:uncertainty}

\begin{figure}
	\centering
	\includegraphics[width=0.98\linewidth]{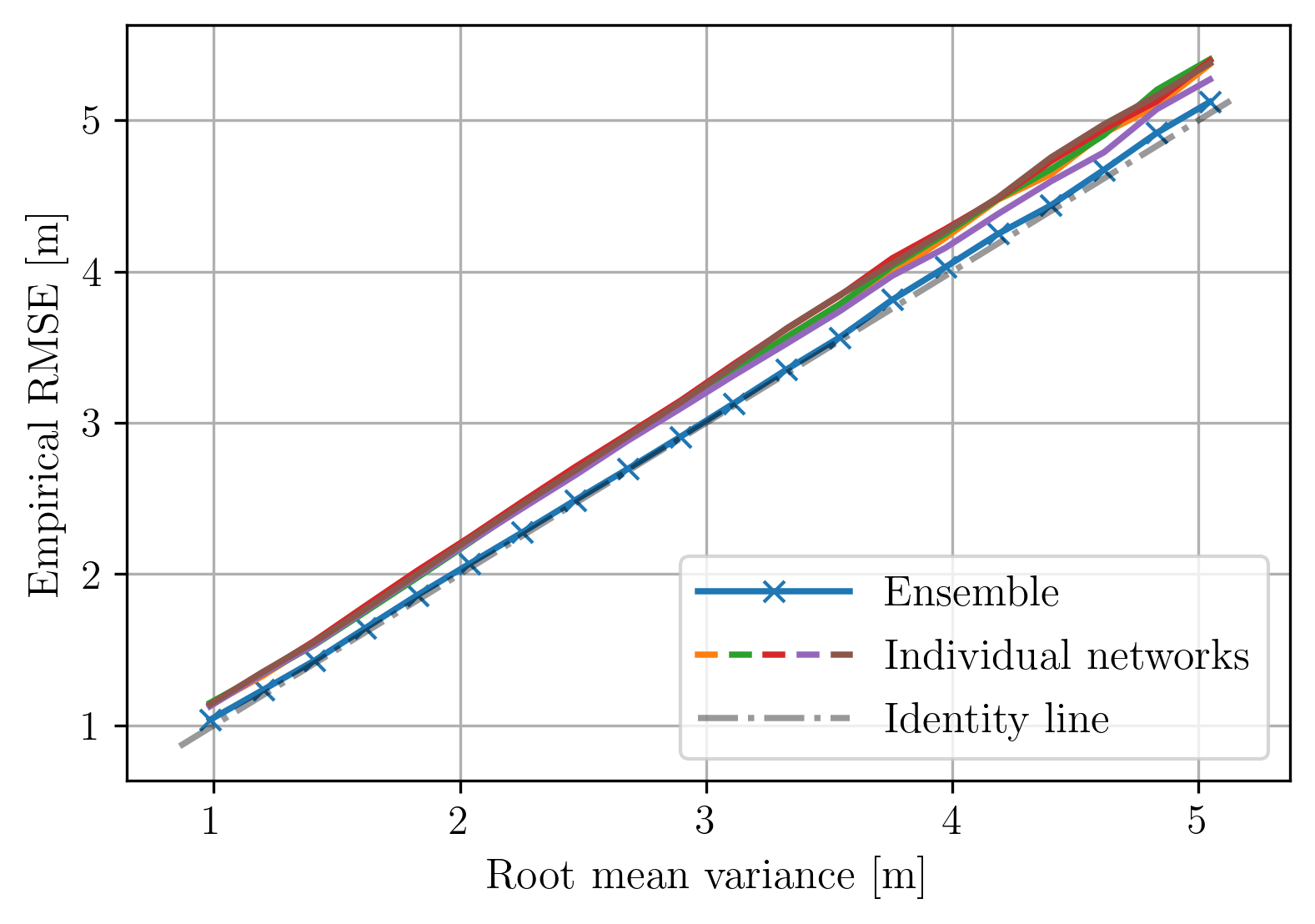}
	\caption{Calibration plot for the \texttt{P95} variable, using 20 uncertainty bins. For the ensemble, the predicted uncertainty agrees very well with the actually observed error, whereas we observe systematic over-confidence in the individual models' predictions. Note that for this plot, we have removed the first and last percentile of predictions (according to predicted uncertainty) to avoid bins with a very small number of samples at the tails of the distribution, because with very few samples the MSE is a noisy estimate of the expected squared error.}
	\label{fig:P95_calibration}
\end{figure}

We evaluate the quality of the model's uncertainty estimates (Eq.~\ref{eq:pred_var}), i.e., how well the predicted uncertainty correlates with the expected error of a given prediction. If a model is perfectly calibrated, then, for any value of the predicted variance, the expected squared error between the true value and the predicted mean should be equal to the predicted variance (by definition of the variance). To obtain an estimate of the expected squared error, we group predictions into bins based on the predicted uncertainty and compare, for each bin, the mean squared error with the mean predicted variance of all samples in the bin. For a more intuitive interpretation, in the units of the original variables, we take square roots and compare the empirical RMSE and the root mean variance.
In Fig.~\ref{fig:P95_calibration}, the result of this comparison is shown for 20 bins for the \texttt{P95} variable. It can be observed that the plot corresponding to the ensemble follows the identity line closely, implying highly reliable uncertainty estimates. Only for higher-variance predictions ($\text{Var}[\bm{y}_* \mid \bm{x}_*,\mathcal{D}] \gtrsim 12$) we can observe a slight under-estimation of uncertainty. Each of the individual neural networks, in contrast to the ensemble, suffers from much more severe under-estimation of uncertainty (i.e., they are over-confident) throughout all uncertainty levels. This is consistent with the literature, where it has repeatedly been observed that neural network models tend to be overconfident about their predictions \citep{Guo2017, Lakshminarayanan2017}. The calibration plots for the remaining structure variables are qualitatively similar and can be found in the appendix (Fig.~\ref{fig:other_calibration}). Note that even though the ensemble reaches the best calibration for all structure variables (compared to the individual neural networks), the individual networks' uncertainty estimates are, in our specific case, already very good. We attribute this to the large and diverse dataset that was available for training, such that only few test samples are insufficiently represented by the training set and require the (approximate) Bayesian marginalization to achieve good model calibration.

\begin{figure}
	\centering
	\includegraphics[width=\linewidth]{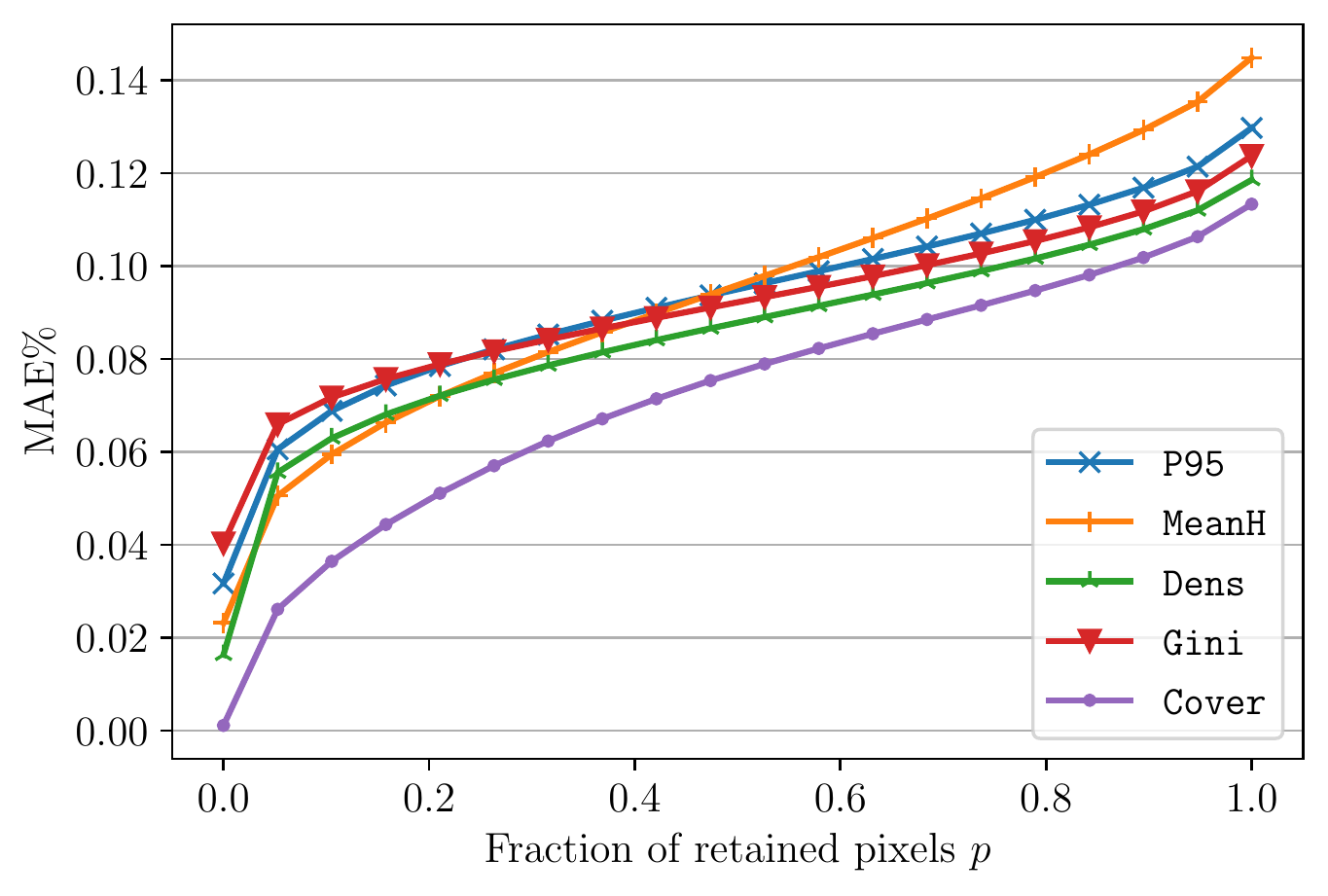}
	\caption{MAE\% for all variables measured on a fraction $p \in (0,1]$ of test pixels that have the lowest associated uncertainty (as defined by Eq.~\ref{eq:pred_var}). The plot shows a strong correlation between estimated uncertainty and expected normalized absolute error. We show the normalized MAE\% (instead of the absolute MAE) so we can compare the results for the five structural variables in a single plot.}
	\label{fig:uncertainty_correlation}
\end{figure}

Another way to validate the usefulness of the predicted uncertainty values is to observe how error metrics change when retaining a decreasing fraction $p \in (0,1]$ of least uncertain predictions. The intuition is that if one discards predictions with high uncertainty and the latter is indeed a good predictor of the expected error, then the average error of the remaining predictions should decrease. Fig.~\ref{fig:uncertainty_correlation} displays this relationship between the fraction of retained pixels and the MAE\%, for all structural variables. Indeed, all curves increase monotonically, implying that the predicted uncertainty correlates (in expectation) with the actual error at a given test sample.

\subsection{Sensor ablation study}
\label{sec:ablation}
We investigate how much the reported accuracy depends on the individual optical and SAR image inputs. To this end, Tab.~\ref{tab:ablation} reports the MAE when training a network with different input data configurations. Unsurprisingly, the default configuration that uses one Sentinel-2 optical and two Sentinel-1 SAR images (``S2+S1") performs best w.r.t.\ all predicted structure variables. What stands out is that removing the optical information (``S1 only") is significantly more detrimental than removing the SAR input (``S2 only"), indicating that the former is more predictive of forest structure.
We assume that this is mainly due to two reasons: (1) Sentinel-1 imagery is noisier, which impairs the analysis of small structures at the level of individual pixels and below; and (2) Sentinel-1 is more heavily affected by topographic influences. In particular, the mountainous topography prevalent in Norway can give rise to SAR-specific effects, such as shadowing, foreshortening and layover \citep{Small1995, Carrasco1997}.

The ablation study further shows that including SAR data from both ascending and descending orbits is beneficial in terms of regression performance, compared to using only one (randomly chosen) orbit direction during training and testing.\footnote{We argue that choosing the orbit direction randomly is the most natural way of disregarding that factor, without introducing biases specific to a given orbit direction} As an explanation, we suspect a combination of two effects: (1) The model has a second observation for any given pixel, thus benefiting from redundancy to suppress noise; and (2) the scene is illuminated from two different directions, which helps to mitigate SAR-specific effects as outlined previously.

Lastly, the ablation study demonstrates that the impact of removing one of the SAR images is far less pronounced when additional optical information is available (``S2+S1" vs.\ ``S2+S1Rand"), compared to a configuration that solely relies on SAR (``S1 only" vs.\ ``S1Rand only"). This outcome further supports the finding that optical images are the more predictive data source for forest structure.

\begin{table}
	\begin{center}
	    \caption{MAE of the reported structural variables for various input configurations. The version that uses both Sentinel-2 optical and Sentinel-1 SAR imagery performs best, while leaving out the optical input is more harmful compared to leaving out the SAR. The experiments also demonstrate the effectiveness of using SAR images from both ascending and descending orbits, as compared to only one, random orbit direction (``S1Rand").}
		\begin{tabular}{@{}l|c|c|c|c|c@{}}
			Input & \texttt{P95} & \texttt{MeanH} & \texttt{Dens} & \texttt{Gini} & \texttt{Cover} \\
			\hline
			S2+S1 & \textbf{1.697} & \textbf{1.162} & \textbf{0.063} & \textbf{0.029} & \textbf{0.080} \\
			S2+S1Rand & 1.778 & 1.219 & 0.066 & 0.029 & 0.083 \\
			S2 only & 1.813 & 1.243 & 0.067 & 0.029 & 0.084 \\
			S1 only & 3.052 & 2.123 & 0.120 & 0.038 & 0.148 \\
			S1Rand only & 3.480 & 2.421 & 0.144 & 0.040 & 0.174 \\
		\end{tabular}
		\label{tab:ablation}
	\end{center}
\end{table}

\subsection{Country-wide forest structure map}
To demonstrate the applicability of our method at country-scale, we compute a Norway-wide map of forest structure variables for the year 2020 and make it publicly available. To be more robust against clouds and to avoid gaps, we use $T=10$ optical images per location that were acquired throughout the leaf-on-period June to October. We also add one SAR image to every location and pair it with the optical data. We feed all eleven inputs per location to our model, which consists of five independently trained neural networks as described in Sec.~\ref{sec:method}. This ultimately yields ten predictions for mean and variance for every geographical location. To produce an easy-to-interpret map, we collapse the ten predictive distributions into a single point estimate by applying an inverse-variance weighting:

\begin{align}
y_{ij} = \frac{\sum_{t=1}^{T} \bar{\mu}_{tij} / \bar{\sigma}_{tij}^2}{\sum_{t=1}^{T} 1 / \bar{\sigma}_{tij}^2},
\label{eq:inverse_var}
\end{align}
for every pixel $i$ and structural variable $j$.

This scheme poses a natural way of combining multiple measurements into a single forest structure map, exploiting the (previously demonstrated) property of our model to return well-calibrated uncertainties. Intuitively, predictions that are assigned a high uncertainty by our model should have less influence on the final results compared to predictions of high confidence. It can be shown that the resulting, inverse-variance weighted forest structure estimates have the smallest expected error among all possible weighted averages \citep{Strutz16}. Therefore, by means of outputting well-calibrated predictions, we are able to combine multiple measurements in an theoretically optimal way, which is not possible when only point estimates are available.

\begin{figure*}[ht!]
	\centering
	\includegraphics[width=0.32\linewidth,trim={24px 24px 24px 24px},clip]{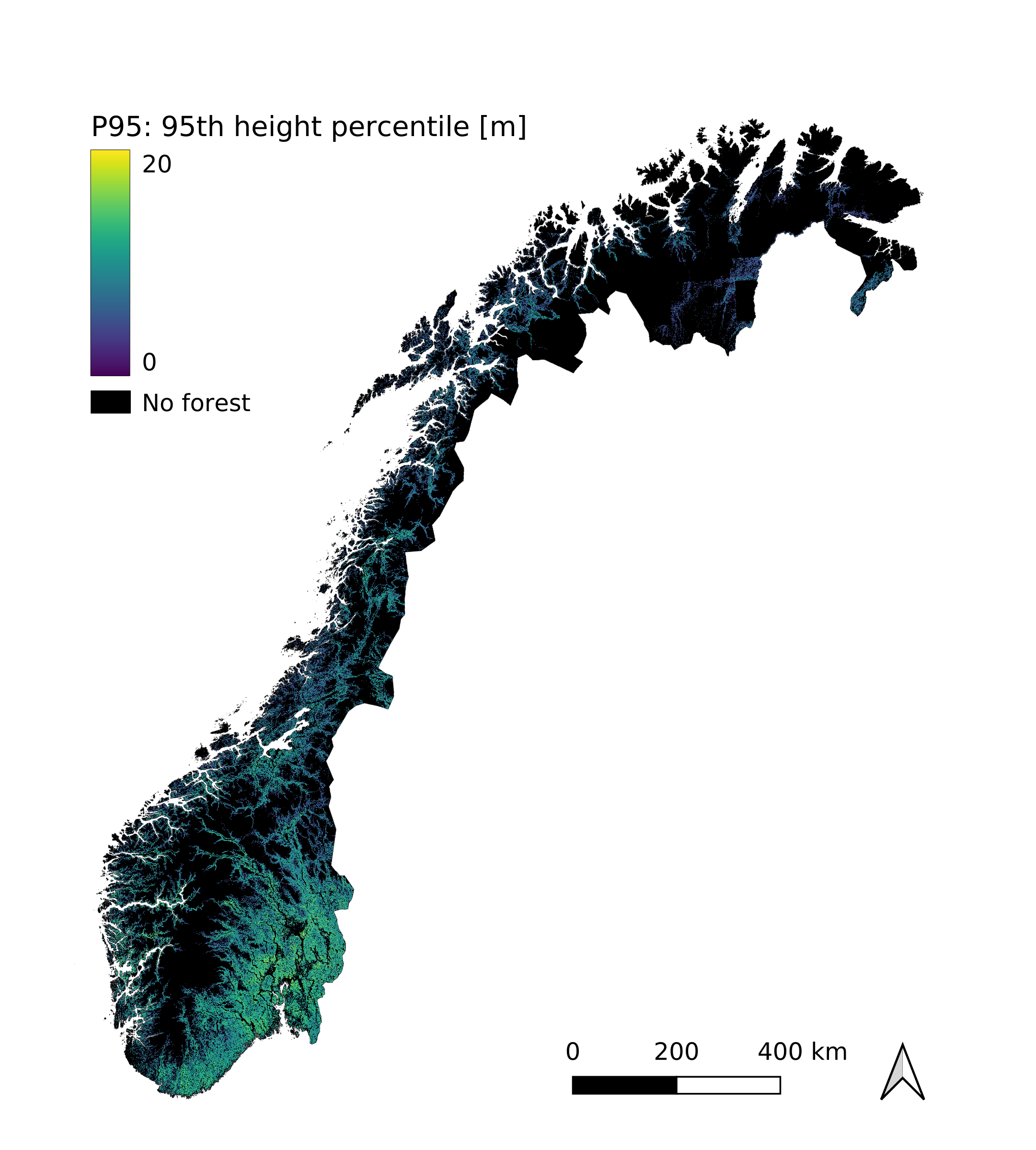}
	\includegraphics[width=0.32\linewidth,trim={24px 24px 24px 24px},clip]{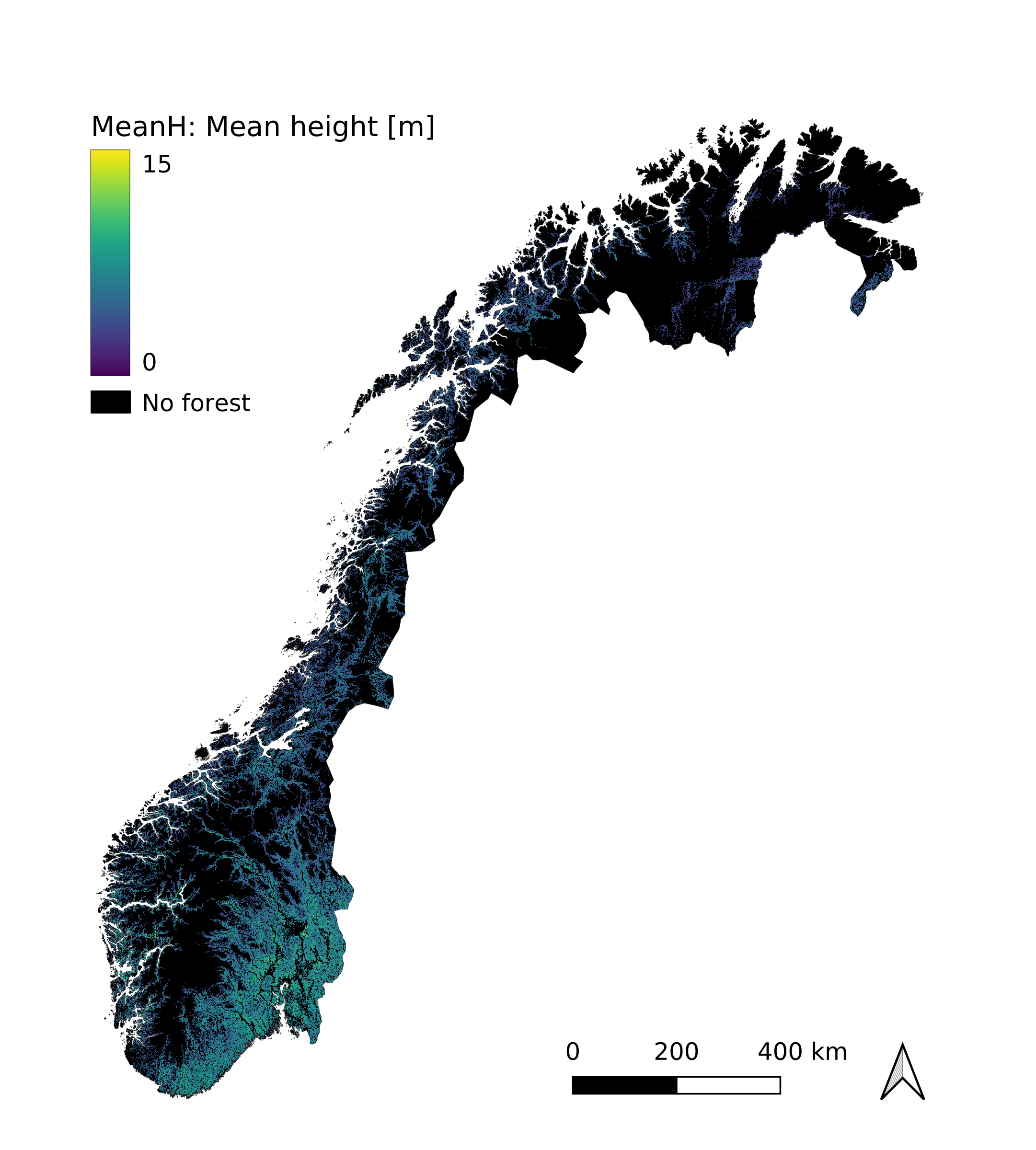}
	\includegraphics[width=0.32\linewidth,trim={24px 24px 24px 24px},clip]{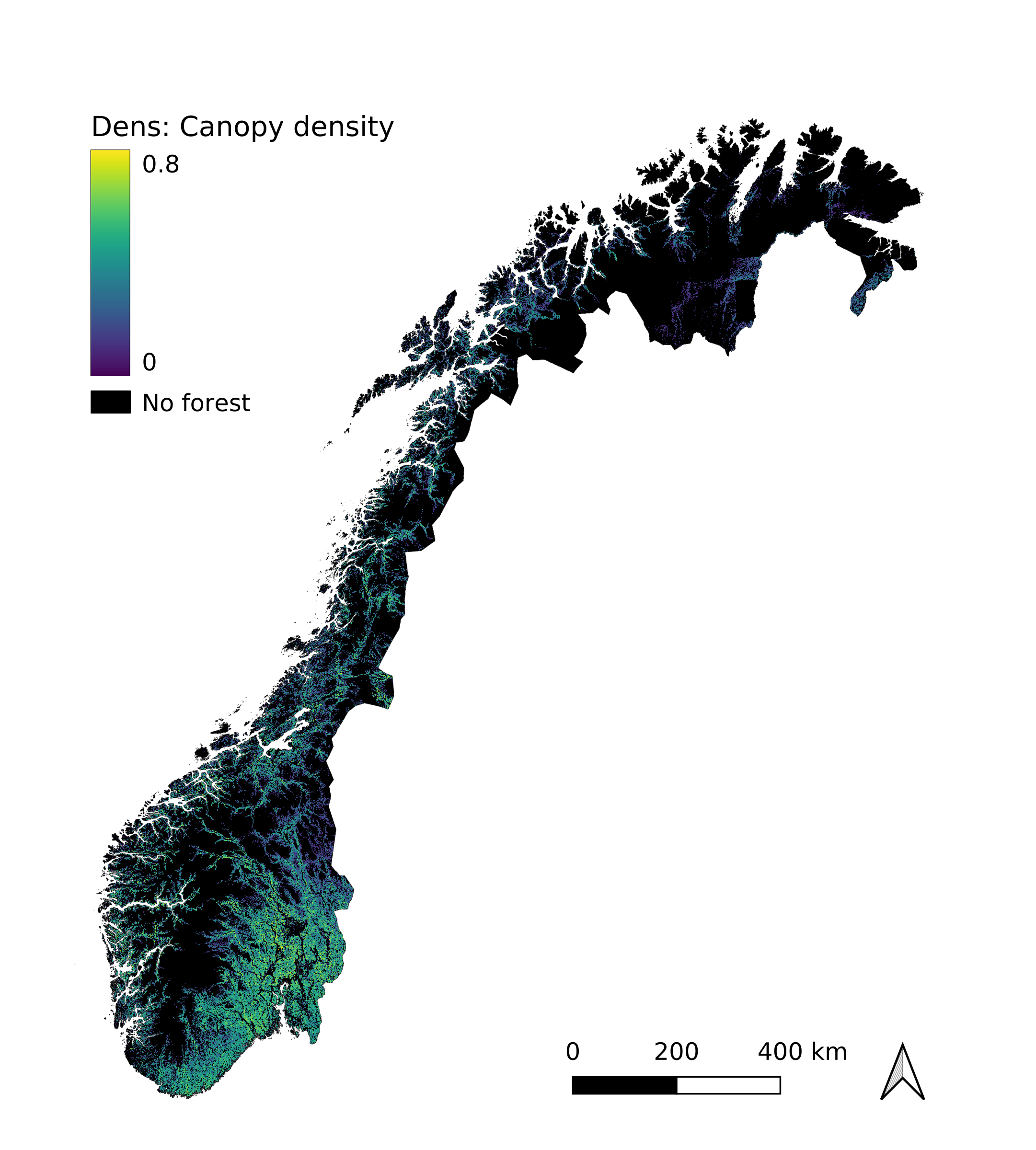}\\
	\vspace{2em}
    \includegraphics[width=0.32\linewidth,trim={24px 24px 24px 24px},clip]{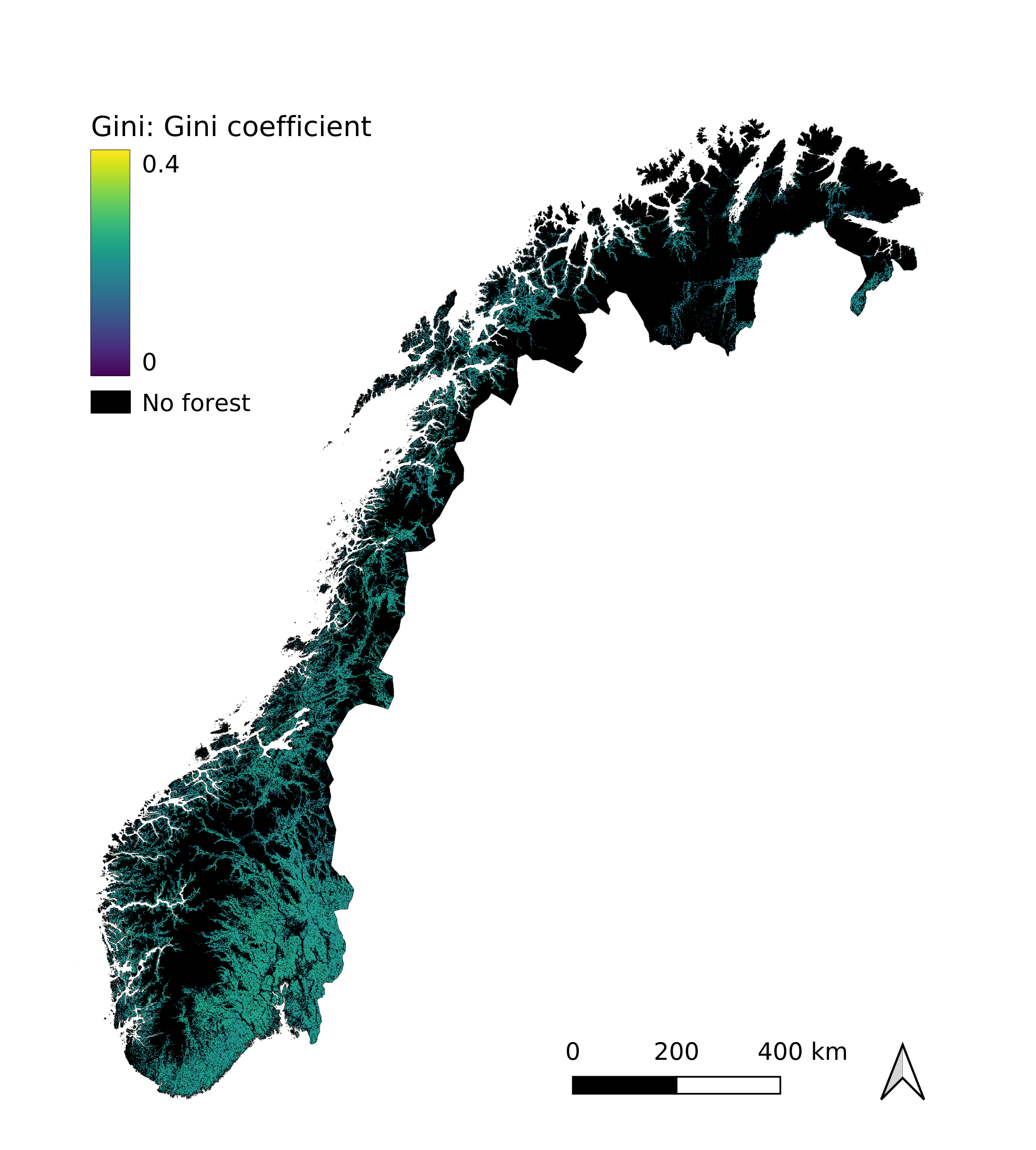}
	\includegraphics[width=0.32\linewidth,trim={24px 24px 24px 24px},clip]{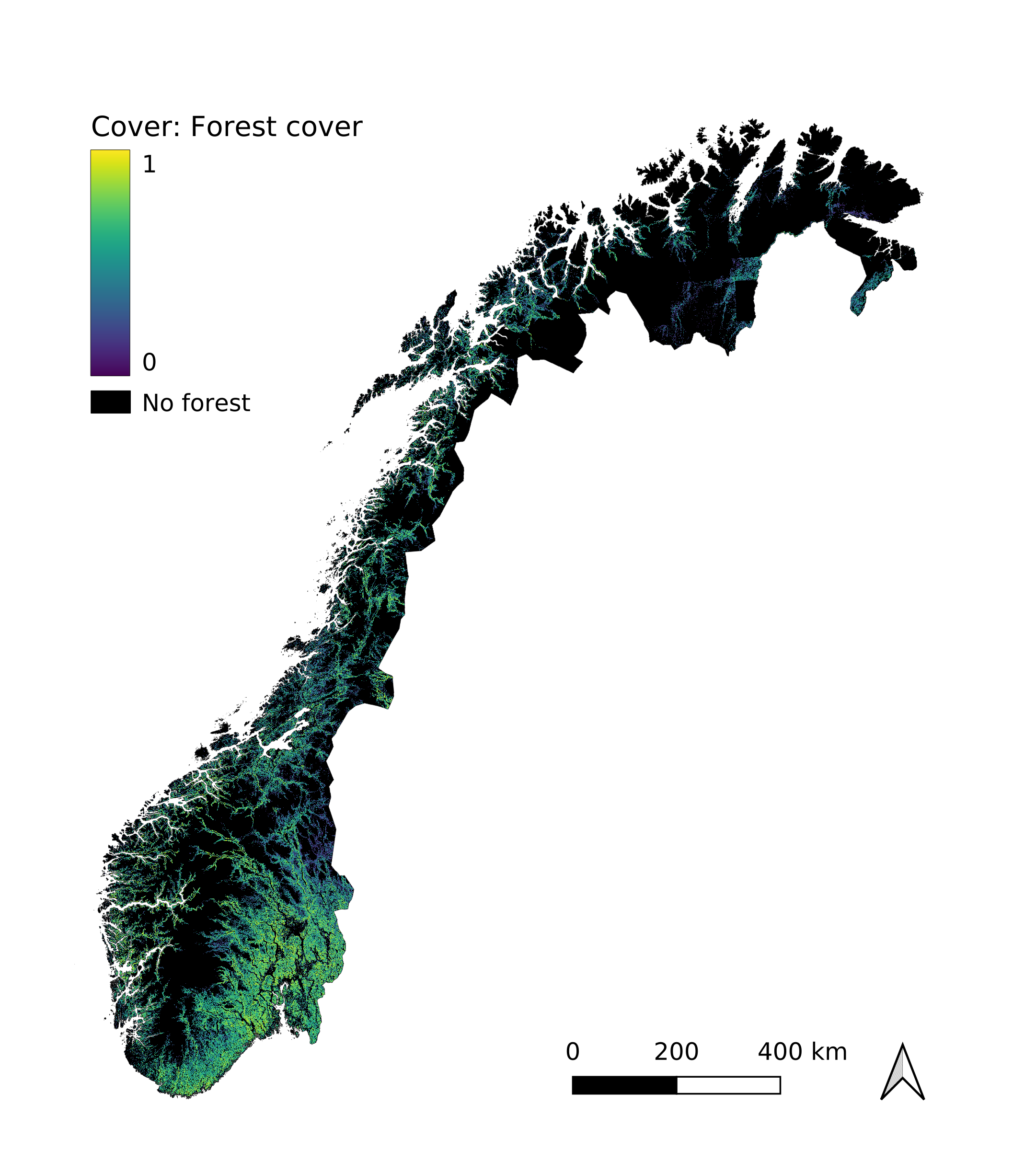}
	\caption{Overview images of the five channels of the produced Norway-wide forest structure map, corresponding to the five structural variables. Non-forested areas have been masked out and labeled as ``No forest".}
	\label{fig:map}
\end{figure*}

Fig. \ref{fig:map} shows overview images of the generated map for each of the five structural forest variables. We mask out areas that are not considered forested according to the NIBIO timber volume map (see Sec.~\ref{sec:model_training}) as already done for training our model\footnote{Note that two small regions in northern Norway are missing in our map due to missing data in the forest mask.}.

\section{Conclusion}
\label{sec:conclusion}

The main finding of our research is that forest structure indicators can be mapped at country-scale from publicly available Sentinel data, in combination with modern deep machine learning and open airborne laser scanning (ALS) data as reference data for training. The availability of these maps at almost any point in time (starting from the beginning of the Copernicus program in 2015) opens up new possibilities for large scale forest monitoring and resource mapping and is thus a promising tool for understanding and tackling challenges such as climate change or biodiversity loss. A straightforward application of the predicted structural variables could be updating existing nationwide forest resource maps hitherto produced on the basis of the same variables, but derived from ALS data \citep{Nord-Larsen2012,Nilsson2017,Astrup2019}. Given the large costs for updating these maps, our Sentinel-derived forest structure variables could provide inexpensive auxiliary data for producing dense time series of forest resource maps. The availability of such dynamic maps can help to better understand the effect of forest management practices on forest biomass dynamics and on the variations in functional diversity. In addition to their use for mapping traditional variables (e.g., above-ground biomass), dense time series of canopy height (e.g., \texttt{P95}) from Sentinel data allow one to derive the canopy height vertical growth, which in turn can be used to measure site productivity \citep{Noordermeer_2018_siteIndex,Solberg_age_indep_2019}. In addition, the produced maps can also be used to provide proxies to identify and map forest disturbances on a large scale  \citep{Hansen2013, Senf2018}. 

An innovative aspect of our method is that it additionally predicts variables describing the vegetation density (\texttt{Dens}) and its variation in the vertical profile of the canopy (\texttt{Gini}), different to previous studies that focused on canopy height or cover only. These additional variables are important as they provide complementary information to the height and the cover and thus allow for a more comprehensive understanding of forest structure on a large scale. The combined use all of these variables further opens up new possibilities to address complex issues, such as the quantification of biomass losses and gains from subtle land-use changes, e.g., forest degradation and forest restoration. 

A further novelty of the presented method, compared to previous applications of deep learning to remotely sensed data, is that it outputs pixel-wise, well-calibrated output distributions instead of mere point estimates. The quantification of the predictive uncertainty makes the system more reliable and more trustworthy. In particular, estimates of the expected predictive error can be propagated to downstream tasks, such as higher-level mapping systems that rely on our forest structure estimates as input. Even further downstream, maps ultimately serve as a basis for forest management, where well-calibrated uncertainty estimates (``knowing what we don't know") translate to better decision making.

\bibliographystyle{elsarticle-num-names}\biboptions{authoryear}
\bibliography{cas-refs}

\clearpage

\appendix

\section{Derivations}

In the following, we provide mathematical derivations for the employed loss function (Eq.~\ref{eq:loss}) and the moments of the posterior predictive distribution (Eqs.~\ref{eq:pred_mean} and \ref{eq:pred_var}).
\subsection{Maximization of the parameter posterior}
\label{ap:loss}
A standard approach in machine learning (see e.g. \citet{Goodfellow2016}), we start with the maximizer of the logarithmic posterior parameter probability (i.e., its mode) and then deduce the equivalence to the loss function presented in Eq.~\ref{eq:loss}:

\begin{subequations}
\begin{align}
&\argmax_{\bm{\theta}} \log p(\bm{\theta} \mid \mathcal{D}) \\
&=\argmax_{\bm{\theta}} \log \underbrace{p(\bm{\theta})}_{\text{prior}} + \log \underbrace{p(\mathcal{D} \mid \bm{\theta})}_{\text{likelihood}} - \log \underbrace{p(\mathcal{D})}_{\text{evidence}} \label{eq:loss_deriv_2} \\
&= \argmax_{\bm{\theta}} \log p(\bm{\theta}) + \log p(\mathcal{D} \mid \bm{\theta}) \label{eq:loss_deriv_3} \\
&= \argmin_{\bm{\theta}} -\log p(\bm{\theta})-\sum_{i=1}^{N} \log p(\bm{y}_{i} \mid \bm{x}_{i}, \bm{\theta}) \label{eq:loss_deriv_4}\\
&= \argmin_{\bm{\theta}} -\log \mathcal{N}(\bm{\theta}; \bm{0}, \sigma_p^2 I) - \sum_{i=1}^{N} \log \mathcal{N}(\bm{y}_{i}; \hat{\bm{\mu}}_{i}, \hat{\Sigma}_i) \label{eq:loss_deriv_5} \\
&= \argmin_{\bm{\theta}} \frac{\lambda}{2} \left\|\bm{\theta}\right\|_2^2 + \frac{1}{2} \sum_{i=1}^{N}  (\bm{y}_{i}-\hat{\bm{\mu}}_{i})^T \hat{\Sigma}_i^{-1} (\bm{y}_{i}-\hat{\bm{\mu}}_{i}) \label{eq:loss_deriv_6} \\
&= \argmin_{\bm{\theta}} \underbrace{\lambda \left\|\bm{\theta}\right\|_2^2 + \sum_{i,j} \big[ \hat{s}_{ij} + \exp(-\hat{s}_{ij}) (\hat{\mu}_{ij} - y_{ij})^2 \big]}_{\mathcal{L}(\mathcal{D}; \bm{\theta})}. \label{eq:loss_deriv_7}
\end{align}
\end{subequations}
In Eqs.~\ref{eq:loss_deriv_2} and \ref{eq:loss_deriv_3}, we apply Bayes' rule (in log space) and then drop the evidence term as it does not depend on $\bm{\theta}$.
We assume independent and identically distributed labels given the respective input and the model parameters and can thus factor the likelihood into individual data point likelihoods (Eq.~\ref{eq:loss_deriv_4}).
In Eq.~\ref{eq:loss_deriv_5}, $\mathcal{N}(\cdot\mid \bm{\mu}, \Sigma)$ denotes the probability density function of the multivariate Gaussian normal distribution with mean $\bm{\mu}$ and covariance $\Sigma$. We assume an isotropic Gaussian with variance $\sigma_p^2$ as prior over the parameters, as it is common in machine learning (this is equivalent to $\mathcal{L}_2$ regularization). 
In the same line, $\hat{\Sigma}_i = \text{diag}(\hat{\bm{\sigma}}_i^2)$ is a diagonal matrix containing the predicted variances for all structural variables as diagonal elements.
To arrive at Eq.~\ref{eq:loss_deriv_6}, we plug in the Gaussian density function and observe that $\lambda \propto 1 / \sigma_p^2$ governs the strength of regularization.
In Eq.~\ref{eq:loss_deriv_7}, $j \in \{1\dots5\}$ indexes the structure variables and we again use $\hat{s}_{ij} = \log \hat{\sigma}_{ij}^2$.

\subsection{Approximate Bayesian marginalization}
The posterior predictive distribution $p(\bm{y}_* \mid \bm{x}_*,\mathcal{D})$ is obtained by marginalizing over the posterior parameter distribution $p(\bm{\theta} \mid \mathcal{D})$:
\begin{subequations}
\begin{align}
&p(\bm{y}_* \mid \bm{x}_*,\mathcal{D}) \\
&= \int p(\bm{y}_* \mid \bm{x}_*, \bm{\theta}) \; p(\bm{\theta} \mid \mathcal{D}) \; d \bm{\theta} \label{eq:predictive_deriv_1}\\
&= \mathbb{E}_{\bm{\theta} \sim p(\bm{\theta} \mid \mathcal{D})} \; \big[ p(\bm{y}_* \mid \bm{x}_*, \bm{\theta}) \big] \\
&\approx \frac{1}{M} \sum_{k=1}^M \mathcal{N}\big(\bm{y}_* \mid \hat{\bm{\mu}}_{*,k}, \text{diag}(\hat{\bm{\sigma}}^2_{*,k})\big)\;,
\label{eq:predictive_deriv_3}
\end{align}
\end{subequations}
where $\hat{\bm{\mu}}_{*,k} := \hat{\bm{\mu}}(\bm{x}_{*};\bm{\theta}_k)$ and $\hat{\bm{\sigma}}^2_{*,k} := \hat{\bm{\sigma}}^2(\bm{x}_{*};\bm{\theta}_k)$ and $\bm{\theta}_k \sim p(\bm{\theta} \mid \mathcal{D})$. 
Due to the intractability of the integral in Eq.~\ref{eq:predictive_deriv_1}, we estimate this quantity using Monte Carlo sampling (Eq.~\ref{eq:predictive_deriv_3}).
In practise, we sample multiple $\bm{\theta}_k$ by training an ensemble and treating each neural network as a sample from an approximate $p(\bm{\theta} \mid \mathcal{D})$ \citep{Gustafsson2020, Wilson2020}. Given the resulting approximate predictive distribution (Eq.~\ref{eq:predictive_deriv_3}), which is a mixture of Gaussians, the mean is obtained by the tower rule,
\begin{align}
\mathbb{E}[\bm{y}_* \mid \bm{x}_*,\mathcal{D}] = \underset{\bm{\theta} \mid \mathcal{D}}{\mathbb{E}} \Big[ \mathbb{E}[\bm{y}_* \mid \bm{\theta}, \bm{x}_*] \Big] \approx \frac{1}{M} \sum_{k=1}^M \hat{\bm{\mu}}_{*,k},
\end{align}
and the variance by the law of total variance:
\begin{subequations}
\begin{align}
&\text{Var}[\bm{y}_* \mid \bm{x}_*,\mathcal{D}] \\
&= \underset{\bm{\theta} \mid \mathcal{D}}{\mathbb{E}} \Big[\text{Var}[\bm{y}_* \mid \bm{\theta}, \bm{x}_*]\Big] + \underset{\bm{\theta} \mid \mathcal{D}}{\text{Var}}\Big[ \mathbb{E}[\bm{y}_* \mid \bm{\theta}, \bm{x}_*] \Big] \\
&\approx \frac{1}{M} \sum_{k=1}^M \hat{\bm{\sigma}}^2_{*,k} + \frac{1}{M} \sum_{k=1}^M (\hat{\bm{\mu}}_{*,k} - \bar{\bm{\mu}}_*)^2 \\
&= \frac{1}{M} \sum_{k=1}^M [ \hat{\bm{\sigma}}^2_{*,k} + (\hat{\bm{\mu}}_{*,k} - \bar{\bm{\mu}}_*)^2 ]\;,
\end{align}
\end{subequations}
where we again abbreviate $\bar{\bm{\mu}}_* := \sum_{k=1}^M \hat{\bm{\mu}}_{*,k}$.

\section{Additional Experimental Results}

\textbf{Further ablation study results.} Supplementary to Tab.~\ref{tab:ablation}, where we reported the MAE of our method for various input configurations, Tab.~\ref{tab:ablation_all} show the other error metrics for these experiments.\\

\begin{figure}
	\centering
	\vspace{-9em}
	\includegraphics[width=0.7\linewidth]{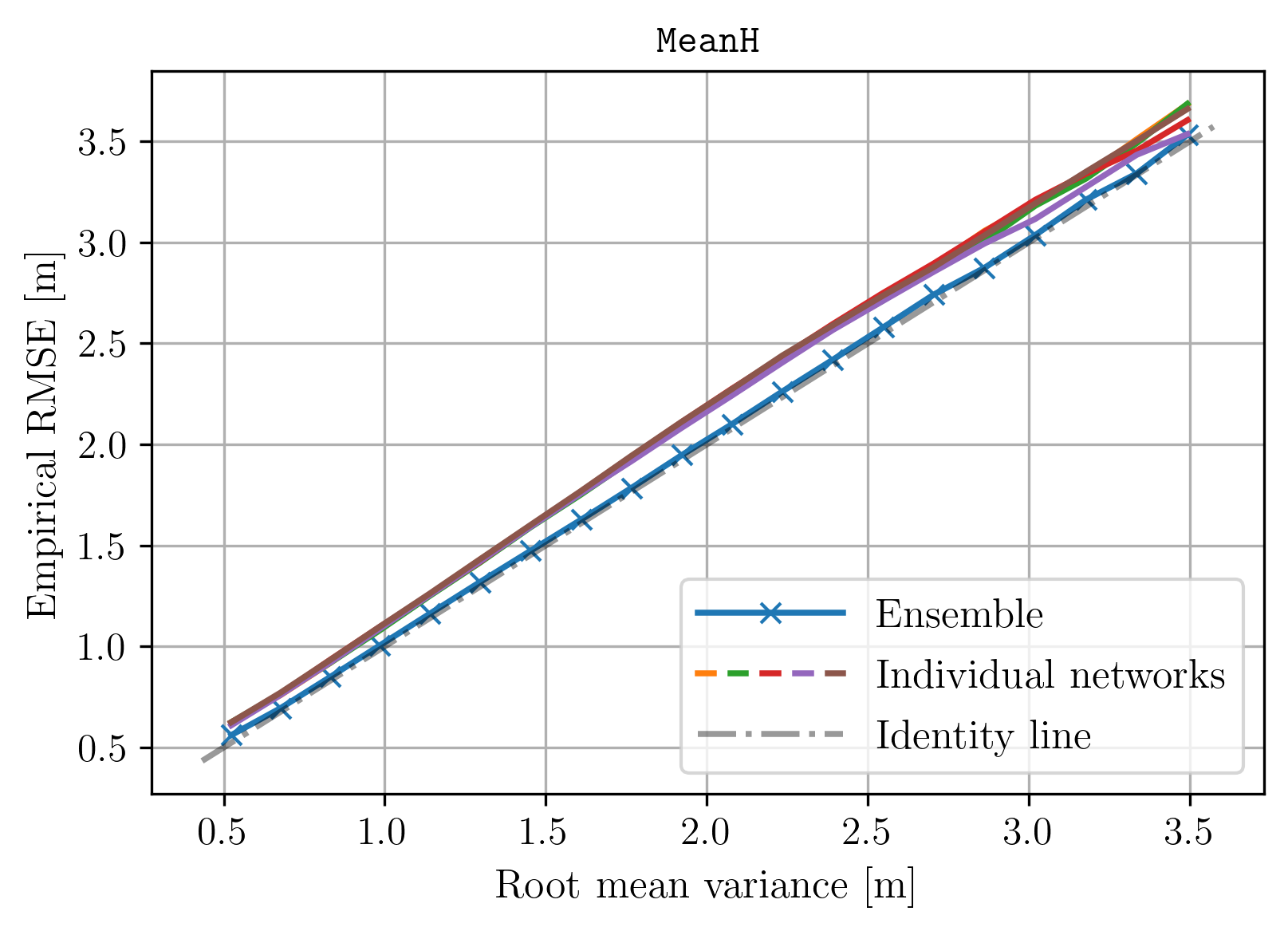}\\
	\includegraphics[width=0.7\linewidth]{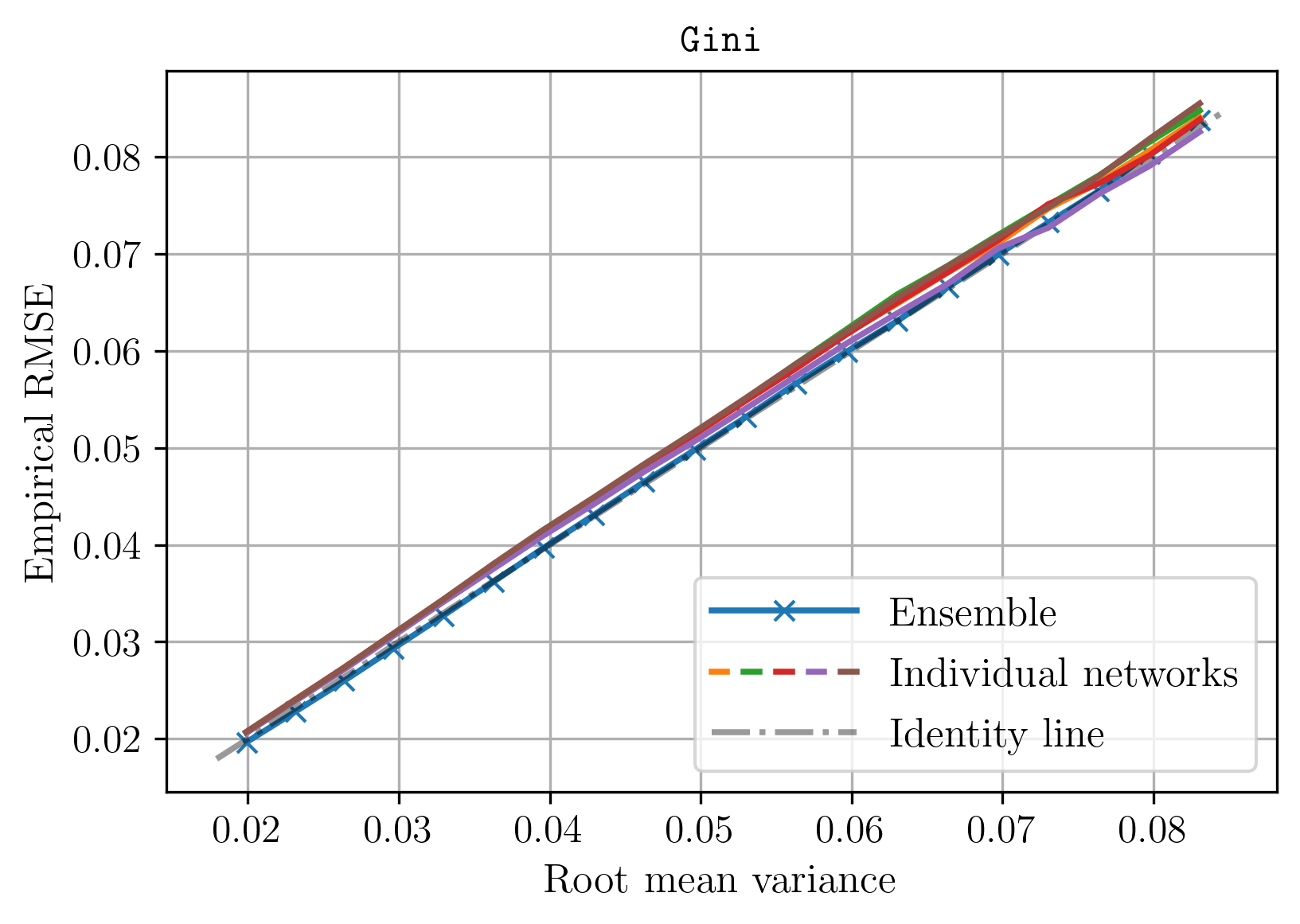}\\
	\includegraphics[width=0.7\linewidth]{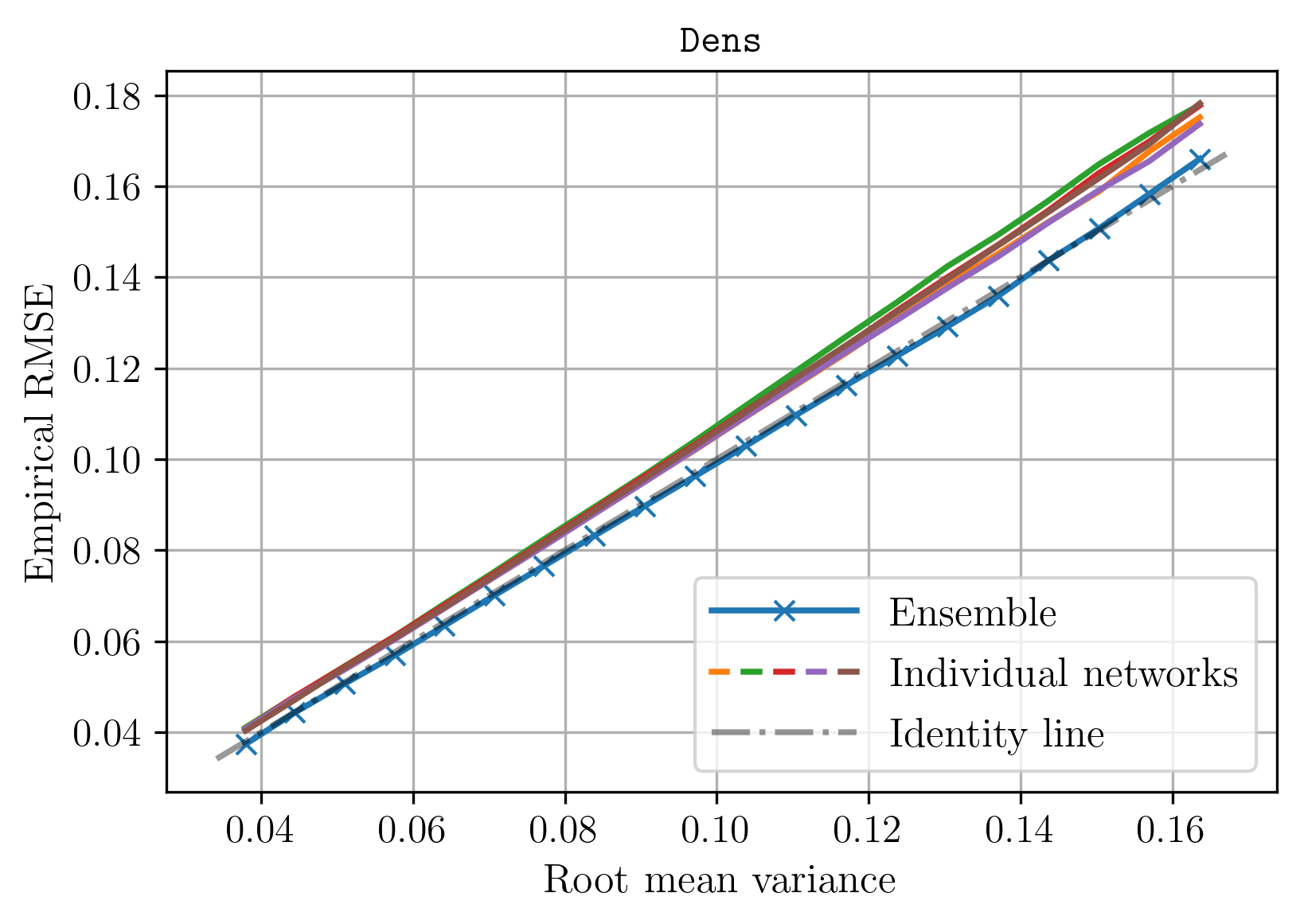}\\
	\includegraphics[width=0.7\linewidth]{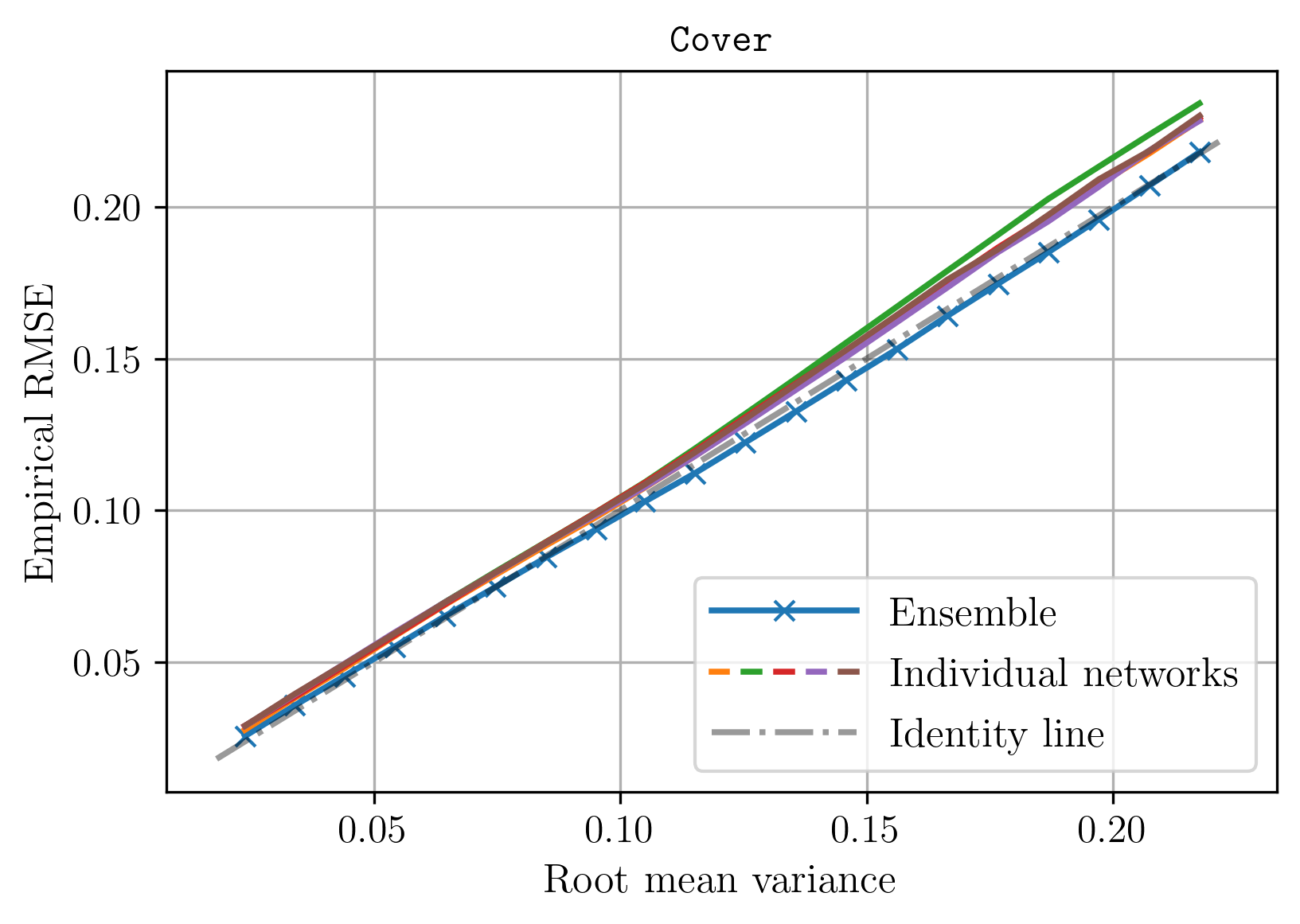}
	\caption{Calibration plots for the \texttt{MeanH}, \texttt{Gini}, \texttt{Dens} and \texttt{Cover} variables, similar to those in Fig.~\ref{fig:P95_calibration}.}
	\label{fig:other_calibration}
\end{figure}

\textbf{Further calibration plots.} Fig.~\ref{fig:other_calibration} shows the calibration plots similar to Fig.~\ref{fig:P95_calibration} for the remaining structure variables. As for \texttt{P95}, we observe a systematic under-estimation of the variance for the individual networks, and a clearly better uncertainty calibration for the ensemble. The effect is more or less pronounced, depending on the variable. For all plots we have again used 20 uncertainty bins and the same noise removal technique as explained for Fig.~\ref{fig:P95_calibration}.\\

\textbf{Baseline comparison.} At the time of writing, we are not aware of any comparable method that predicts diverse forest structure variables from optical and SAR images, and that we could compare to. As a baseline to better understand how our method fares in comparison to a common ``best practice" method for vegetation mapping, we train a random forest (RF) regressor on our data, using per-pixel spectral intensities as inputs. Because of the size of the training dataset ($\approx$65 million data points), we resort to bootstrapping with a sample ratio of $0.05$ and $20$ ensemble members. Tab.~\ref{tab:baseline} compares the test set results of the RF regressor to those of our method. In terms of both MAE and RMSE, the proposed neural network outperforms the RF baseline by large margins for all five structure variables, in most cases reducing the deviations from ground truth by $40-50\%$ (except for \texttt{Gini}, where the gains are only $25-30\%$).
The MBEs are generally low, underlining that both methods are unbiased within the expectable measurement accuracy: all relative MBE-values lie within $\pm2\%$. Overall, the experiment confirms the superior predictive power of deep feature extractors, in particular when trained on large data sets. 

\begin{table}
    \centering
    \caption{Detailed experimental results of the input ablation study conducted in Sec.~\ref{sec:ablation}, reporting the performance of our method on different input configurations.}
    \begin{tabular}{@{}p{0.1cm} c|c|c|c|c|c@{}}
        & & \texttt{P95} & \texttt{MeanH} & \texttt{Dens} & \texttt{Gini} & \texttt{Cover} \\
        \hline
        \multirow{6}{*}{\rotatebox[origin=c]{90}{a) S2+S1}} & MAE & 1.697 & 1.162 & 0.063 & 0.029 & 0.080 \\
		& MAE\% & 0.132 & 0.148 & 0.121 & 0.125 & 0.115 \\
		& RMSE & 2.366 & 1.645 & 0.085 & 0.039 & 0.110 \\
		& RMSE\% & 0.185 & 0.210 & 0.163 & 0.172 & 0.158 \\
		& MBE & -0.039 & -0.019 & -0.002 & -0.000 & -0.001 \\
		& MBE\% & -0.003 & -0.002 & -0.004 & -0.002 & -0.001 \\
        \hline
        \hline
        \multirow{6}{*}{\rotatebox[origin=c]{90}{b) S2+S1Rand}} & MAE & 1.778 & 1.219 & 0.066 & 0.029 & 0.083 \\
		& MAE\% & 0.139 & 0.156 & 0.126 & 0.128 & 0.119 \\
		& RMSE & 2.470 & 1.723 & 0.089 & 0.040 & 0.114 \\
		& RMSE\% & 0.193 & 0.220 & 0.170 & 0.175 & 0.164 \\
		& MBE & -0.075 & -0.041 & -0.004 & -0.001 & 0.002 \\
		& MBE\% & -0.006 & -0.005 & -0.007 & -0.003 & 0.003 \\
        \hline
        \hline
        \multirow{6}{*}{\rotatebox[origin=c]{90}{c) S2 only}} & MAE & 1.813 & 1.243 & 0.067 & 0.029 & 0.084 \\
		& MAE\% & 0.141 & 0.159 & 0.129 & 0.129 & 0.121 \\
		& RMSE & 2.511 & 1.753 & 0.091 & 0.040 & 0.116 \\
		& RMSE\% & 0.196 & 0.224 & 0.174 & 0.177 & 0.167 \\
		& MBE & -0.037 & -0.006 & -0.000 & -0.001 & 0.005 \\
		& MBE\% & -0.003 & -0.001 & -0.001 & -0.003 & 0.007 \\
        \hline
        \hline
        \multirow{6}{*}{\rotatebox[origin=c]{90}{d) S1 only}} & MAE & 3.052 & 2.123 & 0.120 & 0.038 & 0.148 \\
		& MAE\% & 0.238 & 0.271 & 0.231 & 0.167 & 0.212 \\
		& RMSE & 3.932 & 2.791 & 0.154 & 0.050 & 0.190 \\
		& RMSE\% & 0.307 & 0.356 & 0.295 & 0.219 & 0.272 \\
		& MBE & -0.018 & -0.004 & 0.002 & 0.000 & 0.007 \\
		& MBE\% & -0.001 & -0.001 & 0.004 & 0.000 & 0.010 \\
		\hline
        \hline
        \multirow{6}{*}{\rotatebox[origin=c]{90}{e) S1Rand only}} & MAE & 3.480 & 2.421 & 0.144 & 0.040 & 0.174 \\
		& MAE\% & 0.272 & 0.309 & 0.276 & 0.176 & 0.248 \\
		& RMSE & 4.411 & 3.124 & 0.180 & 0.052 & 0.218 \\
		& RMSE\% & 0.344 & 0.399 & 0.346 & 0.229 & 0.312 \\
		& MBE & -0.066 & -0.054 & 0.001 & 0.000 & 0.006 \\
		& MBE\% & -0.005 & -0.007 & 0.002 & 0.002 & 0.009 \\
    \end{tabular}
	\label{tab:ablation_all}
\end{table}

\begin{figure}
	\centering
	\includegraphics[width=\linewidth]{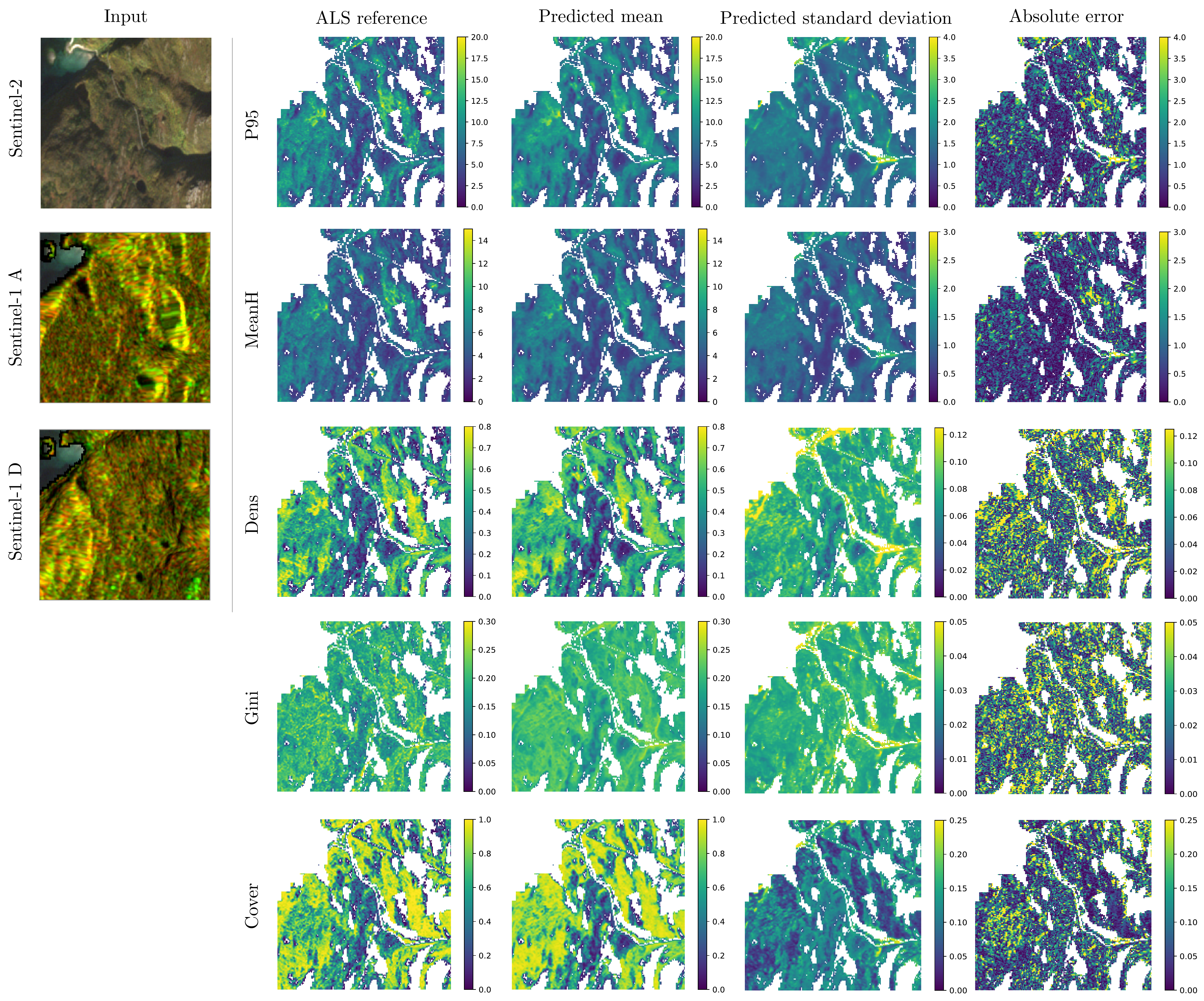}
	\caption{Qualitative prediction example ($180 \times 180$ pixels, 324 ha) from the \emph{North} region, analogous to Fig.~\ref{fig:prediction_sample}.}
	\label{fig:prediction_sample_north}
\end{figure}

\begin{figure}
	\centering
	\includegraphics[width=\linewidth]{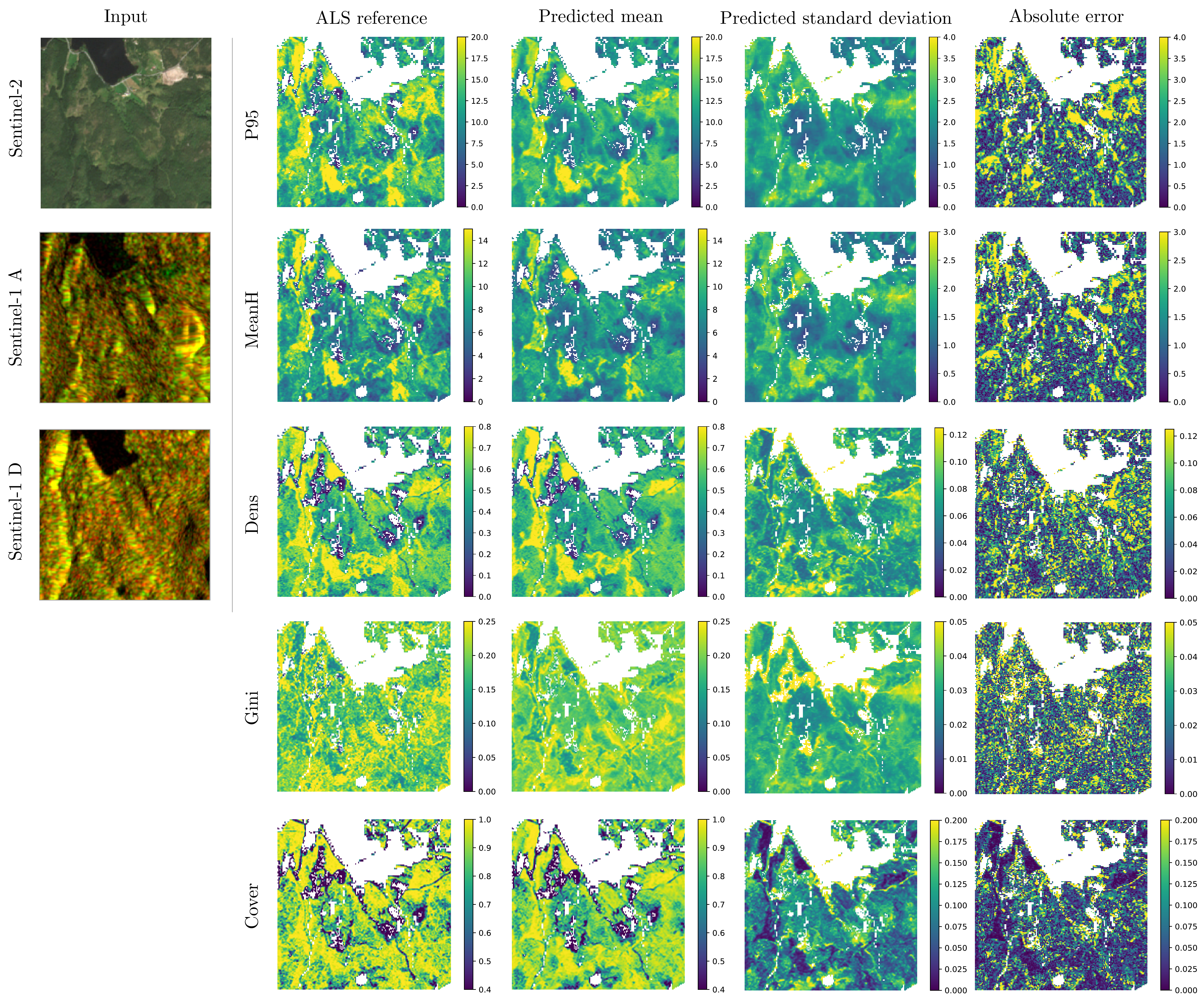}
	\caption{Qualitative prediction sample ($180 \times 180$ pixels, 324 ha) from the \emph{West} region, analogous to Fig.~\ref{fig:prediction_sample}.}
	\label{fig:prediction_sample_west}
\end{figure}

\begin{table}
    \centering
    \caption{Comparison between our proposed method (upper table section, equivalent to the first section of Tab.~\ref{tab:results_all}) and a Random Forest baseline as it is frequently used for the estimation of vegetation parameters from remote sensing images (lower table section). The results indicate that in terms of MAE and RMSE (as well as their normalized counterparts), we outperform the baseline by a large margin for all structural variables.}
    \begin{tabular}{@{}p{0.1cm} c|c|c|c|c|c@{}}
        & & \texttt{P95} & \texttt{MeanH} & \texttt{Dens} & \texttt{Gini} & \texttt{Cover} \\
        \hline
        \multirow{6}{*}{\rotatebox[origin=c]{90}{Ours}} & MAE & 1.648 & 1.127 & 0.061 & 0.028 & 0.078 \\
        & MAE\% & 0.129 & 0.144 & 0.118 & 0.123 & 0.112 \\
        & RMSE & 2.298 & 1.595 & 0.082 & 0.039 & 0.107 \\
        & RMSE\% & 0.179 & 0.204 & 0.158 & 0.170 & 0.154 \\
        & MBE & -0.086 & -0.040 & -0.003 & -0.001 & 0.001 \\
        & MBE\% & -0.007 & -0.005 & -0.006 & -0.004 & 0.002 \\
        \hline
        \hline
        \multirow{6}{*}{\rotatebox[origin=c]{90}{Random Forest}} & MAE & 3.124 & 2.195 & 0.123 & 0.039 & 0.148 \\
		& MAE\% & 0.244 & 0.280 & 0.237 & 0.173 & 0.212 \\
		& RMSE & 4.051 & 2.893 & 0.158 & 0.052 & 0.192 \\
		& RMSE\% & 0.316 & 0.369 & 0.304 & 0.226 & 0.274 \\
		& MBE & 0.111 & 0.095 & -0.002 & -0.001 & 0.000 \\
		& MBE\% & 0.009 & 0.012 & -0.004 & -0.005 & 0.000 \\
    \end{tabular}
	\label{tab:baseline}
\end{table}

\end{document}